\PassOptionsToPackage{table}{xcolor}
\documentclass[11pt]{article}

\usepackage[preprint]{acl}

\usepackage{times}
\usepackage{latexsym}

\usepackage[T1]{fontenc}

\usepackage[utf8]{inputenc}

\usepackage{inconsolata}

\usepackage{graphicx}

\usepackage{subcaption}
\usepackage{booktabs}
\usepackage{multirow}
\usepackage{makecell}
\usepackage{tabularx}
\usepackage{enumitem}
\usepackage{setspace}
\usepackage[most]{tcolorbox}   

\definecolor{bgorange}{RGB}{255, 245, 235}
\definecolor{headerorange}{RGB}{255, 201, 153}
\definecolor{failred}{RGB}{220, 50, 50}
\definecolor{skipblue}{RGB}{0, 102, 204}

\newtcolorbox{failurebox}[1]{
	enhanced,
	colback=bgorange,
	colframe=headerorange,
	arc=6pt,
	boxrule=0.8pt,
	title=\textbf{\textsf{#1}},
	coltitle=black,
	attach boxed title to top left={xshift=5mm, yshift=-3mm},
	boxed title style={colback=headerorange, frame code={}, arc=3pt},
	before skip=10pt,
	after skip=20pt,
}

\definecolor{deepLavender}{rgb}{0.7, 0.5, 0.8}
\usepackage{tikz} 
\newcommand{\mybox}[4]{
	\begin{figure}[ht]
		\centering
		\begin{tikzpicture}
			\node[anchor=text,text width=\columnwidth-0.45cm, draw, rounded corners, line width=1pt, fill=#3, inner sep=2.5mm, align=justify] (big) {\\#4};
			\node[draw, rounded corners, line width=1pt, fill=#2, anchor=west, xshift=5mm] (small) at (big.north west) {#1};
		\end{tikzpicture}
	\end{figure}
}

\newtcbox{\hlprimarytab}{on line, rounded corners, box align=base, colback=green!10,colframe=white,size=fbox,arc=3pt, before upper=\strut, top=-2pt, bottom=-4pt, left=-2pt, right=-2pt, boxrule=0pt}
\newtcbox{\hlsecondarytab}{on line, box align=base, colback=red!10,colframe=white,size=fbox,arc=3pt, before upper=\strut, top=-2pt, bottom=-4pt, left=-2pt, right=-2pt, boxrule=0pt}
\newcommand{\dashifted}{\raisebox{0.5\depth}{\tiny$\downarrow$}}

\newcommand{\da}[1]{{\tiny\hlprimarytab{\dashifted{#1}}}}

\newcommand{\tf}[1]{\textbf{#1}} 

\title{From LLMs to LRMs: Rethinking Pruning for Reasoning-Centric Models}

\author{%
	Longwei Ding $^{1}$ \quad Anhao Zhao$^{1,2}$ Fanghua Ye $^{3}$ \quad Ziyang Chen$^{1,4}$ 
	\quad{Xiaoyu Shen}$^{1}\thanks{Corresponding Author}$\\
	$^1$Institute of Digital Twin, Eastern Institute of Technology, Ningbo  \\
	$^2$Department of Computing, The Hong Kong Polytechnic University\\
	$^3$Tencent Hunyuan / AI Lab \quad 
	$^4$Université Paris Dauphine - PSL 
	\\\small{ \href{mailto:email@domain}{xht102311@gmail.com; xyshen@eitech.edu.cn}
	}
}

\begin{document}
\maketitle

\begin{abstract}
Large language models (LLMs) are increasingly costly to deploy, motivating extensive research on model pruning. However, most existing studies focus on instruction-following LLMs, leaving it unclear whether established pruning strategies transfer to reasoning-augmented models that explicitly generate long intermediate reasoning traces. In this work, we conduct a controlled study of pruning for both instruction-following (\textbf{LLM-instruct}) and reasoning-augmented (\textbf{LLM-think}) models. To isolate the effects of pruning, we align pruning calibration and post-pruning recovery data with each model’s original training distribution, which we show yield more stable and reliable pruning behavior. We evaluate static depth pruning, static width pruning, and dynamic pruning across 17 tasks spanning classification, generation, and reasoning. Our results reveal clear paradigm-dependent differences: depth pruning outperforms width pruning on classification tasks, while width pruning is more robust for generation and reasoning. Moreover, static pruning better preserves reasoning performance, whereas dynamic pruning excels on classification and generation but remains challenging for long-chain reasoning. These findings underscore the need for pruning strategies that explicitly account for the distinct characteristics of reasoning-augmented LLMs. Our code is publicly available at \url{https://github.com/EIT-NLP/LRM-Pruning}.
\end{abstract}
	
\section{Introduction}
Large language models (LLMs) have led to significant advances in natural language processing and enabled impactful applications across a wide range of domains~\cite{multiling_machine_translation, paradigm_shift_in_mt, salute_the_classic, conversational_health_agents, from_llm_to_conversational_agent, evaluating_llm_based_agents_for_multi_turn_conversation}. However, their strong empirical performance comes with substantial computational costs, posing major challenges for practical use in resource-constrained environments \citep{blockpruner,skipgpt}.

\begin{figure*}[ht] 
    \centering
    \includegraphics[width=0.95\textwidth]{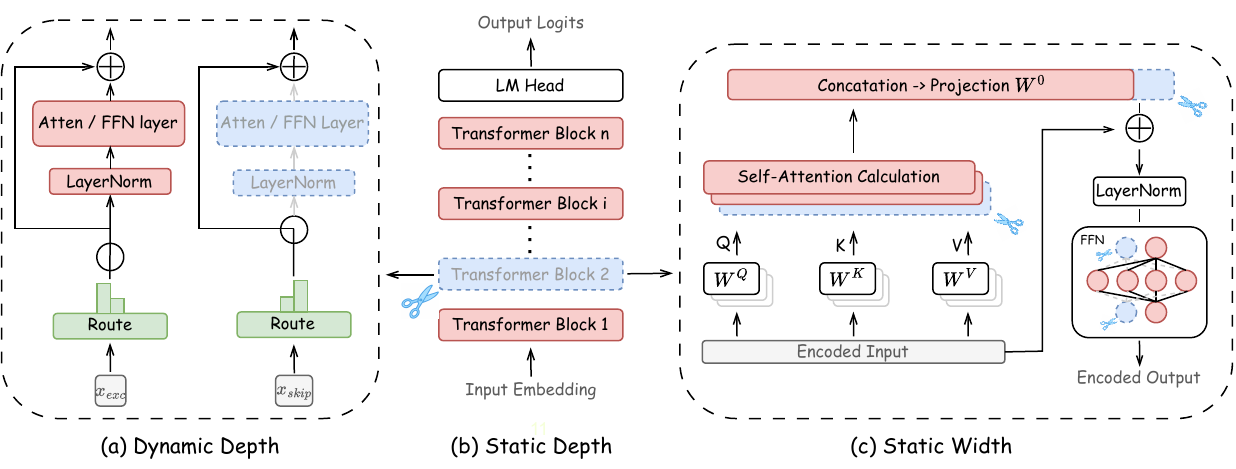} 
    \vspace{-5pt}
    \caption{Overview of the three structured pruning strategies. 
        Static depth pruning removes entire layers, static width pruning reduces hidden dimensions (neurons or attention heads), and dynamic depth pruning adaptively skips layers, attention blocks, or MLP modules depending on the input.
    }
    \label{fig:architecture}
\end{figure*}

Model pruning has emerged as one of the most widely adopted approaches for improving LLM efficiency~\cite{wanda, llm_pruner, shortgpt}. Early work has primarily focused on \emph{static pruning}, where a fixed subset of parameters is removed prior to deployment. Within this paradigm, \emph{static depth pruning} reduces inference cost by removing entire layers, shortening the model and decreasing sequential computation~\citep{shortgpt, shortened_llama}. In contrast, \emph{static width pruning} reduces per-layer computation by shrinking hidden dimensions or pruning channels while preserving the original network depth~\citep{llm_pruner,slicegpt}. More recently, \emph{dynamic pruning} has been explored, which adaptively skips parts of the computation during inference based on input-dependent signals, rather than permanently removing parameters~\citep{mod,jiang2024d,skipgpt}. Together, these techniques form a mature toolkit for accelerating LLMs.

Meanwhile, the LLM landscape is undergoing a paradigm shift from instruction-following models, which we refer to as \textbf{LLM-instruct}, to reasoning-augmented models, denoted as \textbf{LLM-think}~\cite{lrm_survey, deepseek_r1}. Unlike LLM-instruct \citep{qwen2_technical_report}, which generate a direct response with reasoning implicitly embedded, LLM-think explicitly externalize their reasoning process by producing step-by-step intermediate traces prior to the final answer \citep{cot}. While this shift substantially improves performance on complex tasks, it also results in markedly longer generations, often spanning thousands of tokens \citep{do_not_think_that_much, qwen3_technical_report}. Despite these fundamental changes in model behavior and output characteristics, existing pruning research has almost exclusively focused on LLM-instruct. This gap motivates a central question: \textit{does pruning for LLM-think require new strategies, or can existing approaches generalize across paradigms?}

Addressing this question necessitates a carefully controlled pruning pipeline. We therefore begin by revisiting a critical yet often overlooked design choice in prior work: \textit{the selection of datasets used for pruning calibration and post-pruning recovery}. However, prior studies adopt highly heterogeneous datasets for these stages, ranging from general-domain corpora to instruction-tuning data, often without systematic justification \citep{calibration_c4, alpaca}. Such inconsistency introduces confounding factors, making it difficult to disentangle the effects of pruning from those of distribution mismatch.

To enable a principled comparison, we adopt a controlled setting in which calibration and recovery data are strictly aligned with the original training distribution of the target model. Under this setting, we construct matched model–dataset pairs for both paradigms. For \textbf{LLM-instruct}, we use the open-sourced Tulu model~\citep{lambert2024tulu} together with its full instruction-following fine-tuning corpus. For \textbf{LLM-think}, we derive a reasoning-augmented counterpart by further fine-tuning the same base model on OpenThoughts~\citep{guha2025openthoughts}, a dataset designed to elicit explicit multi-step reasoning across diverse difficulty levels. This alignment ensures that each model is pruned, calibrated, and recovered using data drawn from its own training distribution, thereby isolating the impact of pruning itself.
As we show empirically, such alignment best preserves pre-pruning performance and leads to substantially more stable and effective pruning behavior compared to using mismatched datasets. 

Under this controlled setting, we conduct a comprehensive evaluation of three categories of pruning strategies on both LLM-instruct and LLM-think across \textbf{17} diverse classification, generation, and reasoning tasks. Our experiments yield several key empirical findings. First, static depth pruning is more effective than static width pruning on classification tasks, whereas the opposite trend holds for generation and reasoning tasks, where width pruning degrades performance more gracefully. Second, dynamic pruning with unconstrained token-level routing benefits classification and generation but proves substantially harder to optimize for long-chain reasoning, suggesting that aggressive computation skipping is particularly risky for reasoning-intensive workloads. Finally, while dynamic pruning generally outperforms static methods on classification and generation, static pruning retains higher performance on reasoning tasks, highlighting an important gap in current dynamic pruning approaches.

\section{Background}
\label{sec:background}

\paragraph{LLM-instruct and LLM-thinking}
\label{differece_instruct_think}
The rapid progress of LLMs has not only improved their instruction-following ability but also expanded their scope toward explicit reasoning. This shift has given rise to two major families: instruction-following models (\textbf{LLM-instruct}), which are optimized to produce outputs that closely align with user-specified instructions, and reasoning-augmented models (\textbf{LLM-think}), which are trained to generate long chains of thought (CoT) reasoning before arriving at final answers. 
Although the two are built upon the same underlying architecture, recent studies show that they diverge significantly in both internal representations and emergent behaviors.
Mechanistically, LLM-think retain richer contextual information across layers and exhibit higher token-level entropy \citep{layer_by_layer, high_entropy_tokens}. In terms of attention, LLM-instruct generally display diverse specialization across heads, with each attending to different token subsets \citep{fu2024moa}, whereas LLM-think show considerable overlap in tokens attended by different heads \citep{less_is_more}. 

\paragraph{Formalizing Structure Pruning Strategies}
\label{Formalizing_Structure_Pruning_Strategies}
Model pruning in LLMs can be broadly categorized into two approaches: \textit{unstructured pruning}, which removes individual weights based on their magnitudes or importance \citep{unstructured}, and \textit{structured pruning}, which discards entire coupled structures \citep{cheng2024survey}. Although unstructured pruning is conceptually simple, it rarely yields practical acceleration on modern GPUs (e.g., Nvidia GPUs typically require over 90\% sparsity for speedup \citep{nvidia_1,nvidia_2}), while LLMs usually collapse once sparsity exceeds 50\% \citep{song2024sleb}. Therefore, we focus on structured pruning, which is both hardware-friendly and effective. Structured pruning can be further categorized into \textit{width pruning}, which reduces hidden dimensions, and \textit{depth pruning}, which removes redundant layers either statically (the same set of layers is pruned for all inputs) or dynamically (the pruned layers vary depending on the input). The architectures of each method are illustrated in ~\autoref{fig:architecture}. To the best of our knowledge, a unified and systematic formulation of these strategies has not been explicitly articulated in the existing literature. We therefore introduce the following formal definitions, which serve as the foundation for our subsequent analysis. 

Formally, let $\mathbf{H}_l \in \mathbf{R}^{N \times d}$ denote the hidden representations at layer $l$, where $N$ is the sequence length and $d$ is the hidden dimension.

\begin{itemize}[left=0pt]
        \item \textbf{Static width} pruning reduces the hidden dimension $d$ to $d' < d$ by removing less important neurons:
        \begin{equation}
            \mathbf{H}_l' = \mathbf{H}_l[:, \mathcal{I}_l], \quad |\mathcal{I}_l| = d',
            \end{equation}
        where $\mathcal{I}_l$ indexes the retained neurons in layer $l$.
    
        \item \textbf{Static depth} pruning removes entire layers, keeping only a subset $\mathcal{L}' \subseteq \{1, \dots, L\}$ of layers:
            \begin{equation}
                \mathbf{H}_{\mathcal{L}'} = \{\mathbf{H}_l \mid l \in \mathcal{L}'\}, \quad |\mathcal{L}'| = L' < L.
                \end{equation}
        Static depth pruning computes a fixed importance score for each layer and permanently removes those identified as redundant. Typical criteria for scoring include:
            \begin{itemize}[left=0pt]
                    \item \textit{Block Influence (BI)}\citep{shortgpt} measures the cosine similarity between the input and output representations of a layer. Layers with high similarity are considered to contribute little new information and are thus pruned.
        
                    \item \textit{Perplexity (PPL)}\citep{shortened_llama} quantifies the impact of individually removing a layer on validation perplexity. Layers that cause minimal degradation are considered less critical.
        
                    \item \textit{Taylor}\citep{shortened_llama} estimates the sensitivity of the loss to parameter removal through the first-order gradient–weight product, and the layer with low aggregated sensitivity are pruned.  
                \end{itemize}
    
        \item \textbf{Dynamic depth} pruning introduces a router module, denoted as $R(\cdot)$, which determines whether a block is executed or skipped for each input token \citep{raposo2024mixture,jiang2024d,skipgpt}. 
        Let $x$ denote the input to a block, which may correspond to a Transformer layer, an attention module, or a MLP. 
        A binary gate $g$ governs the execution, and is defined as
        \begin{equation}
            g = R(x) \in \{0,1\}.
            \end{equation}
        where $g=1$ indicates execution and $g=0$ indicates skipping. 
        Let $f(\cdot)$ denote the computation performed by the block; the output is then updated as:
        \begin{equation}
            x' = g \cdot f(x) + x.
            \end{equation}
    \end{itemize}

\paragraph{Problem Setting}
\label{Problem_seetting}
Let $M$ and $M'$ denote the original and compressed LLM respectively. 
Following common practice, we quantify the degree of pruning using the \emph{compression ratio}, defined as the reduction in the average number of parameters active per token:~\cite{skipgpt}:

\begin{equation}
    \label{eq:compression_ratio}
    R(M,M') = 1 - \frac{|M'|}{|M|}
\end{equation}
where $|M|$ and $|M'|$ denote the average per-token parameters in the original and pruned models, respectively.
To evaluate utility, we define \emph{performance retention} as the ratio $\text{Perf}(M')/\text{Perf}(M)$, which measures how effectively the pruned model maintains the functional capabilities of the dense baseline. Consequently, our objective is to identify a pruning strategy that maximizes the compression ratio $R(M,M')$ while minimizing degradation in performance retention.

\section{Experimental setup}
\paragraph{Model} 
For LLM-instruct, we adopt the open-source model \texttt{Llama-3.1-Tulu-3-8B-SFT} \citep{lambert2024tulu}, which provides publicly available weights, instruction-tuning data, and detailed training configurations. For LLM-think, we fine-tune \texttt{Llama-3.1-8B-Instruct} on the OpenThoughts dataset \citep{guha2025openthoughts}, which provides diverse, high-quality reasoning examples that elicit multi-step chains of thought.~\footnote{Fine-tuning was performed using Llama-Factory \citep{llamafactory} on $8\times$,H20 (96,GB) GPUs for 3 epochs ($\approx$,488 GPU hours). See \autoref{tab:llmthink_hyperparams} for details.}

Both models share the same \texttt{Llama-3.1-8B} backbone, ensuring a controlled comparison of pruning effects. Crucially, the training datasets for both models are accessible and used during pruning calibration and post-pruning recovery. This ensures that performance recovery is conducted under the native training distribution rather than on downstream task data, providing a consistent and controlled evaluation of pruning strategies.

\paragraph{Evaluation Benchmarks}
To systematically evaluate the instruction-following capabilities of the LLM-instruct and the reasoning capabilities of the LLM-think, we use a diverse suite of \textbf{17} tasks that can be broadly divided into instruction-following and reasoning benchmarks.
For evaluation, each model is assessed on benchmarks aligned with its primary capabilities: LLM-instruct are evaluated on instruction-following benchmarks, whereas LLM-think are evaluated on reasoning benchmarks.

\begin{figure*}[ht]
    \centering
    \includegraphics[width=1\linewidth]{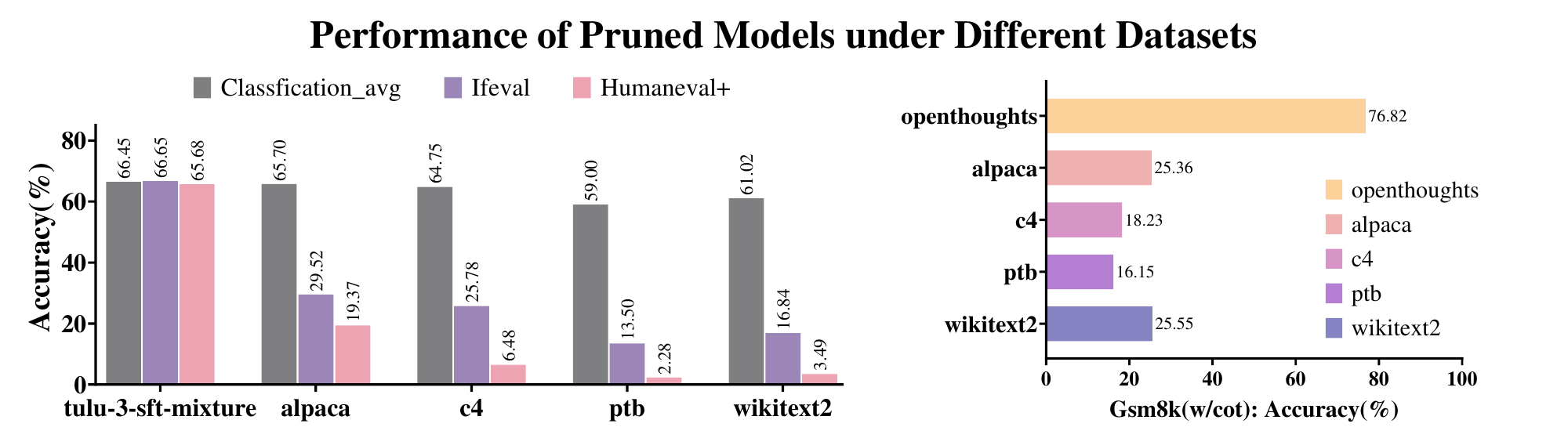}
    \caption{Experimental results validating the impact of calibration and post-training datasets (left: LLM-instruct; right: LLM-think), with results averaged across SliceGPT and ShortGPT.}
    \label{fig:calibration_and_post_trainng}
\end{figure*}

\begin{itemize}[left=0pt]
    \item \textbf{Instruction Following Benchmarks} 
    To comprehensively assess pruned LLM-instruct on instruction following, we include both classification and generation tasks. For \textit{classification} tasks, following \citep{llama}, we include BoolQ \citep{clark2019boolq}, PIQA \citep{piqa}, HellaSwag \citep{zellers2019hellaswag}, WinoGrande \citep{sakaguchi2021winogrande}, ARC-Easy / Challenge (ARC-E / C) \citep{clark2019boolq}, and OpenBookQA (OPQA) \citep{mihaylov2018can}. For \textit{generation} tasks, we adopt a similar setup to \citep{lambert2024tulu}, but exclude tasks that rely heavily on explicit CoT reasoning (evaluated separately under reasoning benchmarks). Specifically, we include IFEval (IFE) \citep{zhou2023instruction} to measure instruction-execution precision; TruthfulQA (TQA) \citep{lin2021truthfulqa} and PopQA (PQA) \citep{mallen2022not} to assess factual accuracy and truthfulness; and HumanEval (HE) \citep{chen2021evaluating} together with HumanEval+ (HE+) \citep{liu2023your} as constrained code-generation tasks, where strict adherence to problem specifications is critical.
    
    \item \textbf{Reasoning Benchmarks} To rigorously evaluate the reasoning abilities of the pruned LLM-think model, we employ five challenging benchmarks spanning mathematical, coding, and scientific domains. Specifically, AIME 2024 (AIME) and MATH-500 (MATH) \citep{lightman2023let} assess advanced mathematical reasoning and multi-step derivation; LiveCodeBench (LCB) \citep{jain2024livecodebench} evaluates code generation, debugging, and comprehension in complex programming tasks; and GPQA-Diamond (GPQA) \citep{rein2024gpqa} together with JEEBench (JEE) \citep{arora2023have} assess nuanced scientific reasoning and the application of domain-specific knowledge.
\end{itemize}

\paragraph{Evaluated Pruning Methods}
To systematically evaluate pruning in both LLM-instruct and LLM-think models, we consider representative methods from three main categories of structured pruning.
\begin{itemize}[left=0pt]
    \item \textbf{Static depth pruning} reduces model depth by removing entire transformer layers. We consider SLEB, an iterative method that recomputes layer-importance scores after each pruning step; \textit{ShortGPT} \citep{shortgpt}, which prunes layers based on input--output representation cosine similarity; and \textit{Shortened-PPL} / \textit{Shortened-Taylor} \citep{shortened_llama}, which identify and remove layers using perplexity (PPL) and Taylor-based criteria, respectively.

    \item \textbf{Static width pruning} reduces model width by pruning attention heads or shrinking the hidden-state dimension. We adopt \textit{LLM-Pruner} \citep{llm_pruner}, a gradient-based method that prunes coupled structures, and \textit{SliceGPT} \citep{slicegpt}, which removes low-variance components from weight matrices via principal component analysis.

    \item \textbf{Dynamic depth pruning} reduces per-input computation by adaptively skipping layers or submodules. \textit{MOD} \citep{raposo2024mixture} selects a subset of tokens to process in each layer via Top-$k$ routing. \textit{D-LLM} \citep{jiang2024d} uses a router to decide whether to execute each transformer layer. \textit{SkipGPT} \citep{skipgpt} combines global token-aware routing with decoupled pruning of MLP and self-attention modules, and defines the pruning ratio as the mean token-level skipping rate across the two modules.
\end{itemize}

\section{Effects of Calibration/Recovery Data}

Before analyzing the effects of different pruning strategies, we first examine a fundamental yet often overlooked design choice in the pruning pipeline:\textit{ the datasets used for calibration and post-pruning recovery}. In modern LLM pruning methods, these two stages constitute the most critical components of the pipeline. Calibration estimates the importance of model structures (e.g., layers or channels) to guide pruning decisions, while post-pruning recovery aims to restore performance by adapting the remaining parameters after pruning. While prior work adopts heterogeneous datasets, their choice for calibration and recovery data is rather arbitrary.

To establish a well-founded experimental setting for subsequent analysis, we first investigate whether aligning both pruning calibration and post-pruning recovery with the model’s original training distribution is beneficial for preserving pre-pruning performance. To this end, we examine two representative pruning methods: ShortGPT for static depth pruning and SliceGPT for static width pruning. We compare general-domain corpora (e.g., C4, PTB, and WikiText2) and standard instruction-tuning datasets (e.g., Alpaca) against the model’s original training corpora, namely \texttt{tulu-3-sft-mixture} for LLM-instruct and \texttt{openthoughts} for LLM-think, under a fixed pruning ratio of 20\%. Additional results that separately disentangle the effects of the calibration dataset and the post-pruning recovery dataset are reported in \autoref{more_detail_cal_pos}.

\begin{table*}[!ht]
    \scriptsize
    \setlength{\tabcolsep}{8pt}
    \renewcommand{\arraystretch}{1}
    \centering
    \begingroup\setlength{\fboxsep}{0.8pt}
    \endgroup
    \begin{tabular}{c|l|ccccccc|c}
        \specialrule{1.5pt}{0pt}{0pt}
        \multirow{2}{*}{\textbf{Ratio}} 
        & \multirow{2}{*}{\textbf{Method}} 
        & \textbf{BoolQ} & \textbf{OBQA} & \textbf{PIQA} 
        & \textbf{WinoGrande} & \textbf{HeSw} & \textbf{ARC-E} 
        & \textbf{ARC-C} & \multirow{2}{*}{\textbf{Avg. Acc.}$\uparrow$} \\
        & & \textit{Acc} & \textit{AccNorm} & \textit{Acc} & \textit{Acc} & \textit{AccNorm} & \textit{AccNorm} & \textit{AccNorm} & \\
        \midrule
        0.00\% & Dense & 82.26 & 46.80 & 80.84 & 77.74 & 82.97 & 87.20 & 61.43 & 74.18 \\
        \cmidrule(lr){1-10}
        
        \multirow{9}{*}{20.0\%} 
        & \cellcolor{gray!15} SLEB              & \cellcolor{gray!15} 77.00 & \cellcolor{gray!15} 39.60 & \cellcolor{gray!15} 75.90 & \cellcolor{gray!15} 73.16 & \cellcolor{gray!15} 74.85 & \cellcolor{gray!15} 78.24 & \cellcolor{gray!15} 48.04 & \cellcolor{gray!15} 66.68 \\
        & \cellcolor{gray!15} ShortGPT              & \cellcolor{gray!15} 68.34 & \cellcolor{gray!15} 39.60 & \cellcolor{gray!15} 74.59 & \cellcolor{gray!15} \underline{75.84} & \cellcolor{gray!15} 75.16 & \cellcolor{gray!15} 79.41 & \cellcolor{gray!15} 51.53 & \cellcolor{gray!15} 66.35 \\
        & \cellcolor{gray!15} Shortened-PPL         & \cellcolor{gray!15} 68.16 & \cellcolor{gray!15} \underline{43.20} & \cellcolor{gray!15} \textbf{78.12} & \cellcolor{gray!15} 64.56 & \cellcolor{gray!15} 71.01 & \cellcolor{gray!15} 78.28 & \cellcolor{gray!15} 48.12 & \cellcolor{gray!15} 64.49 \\
        & \cellcolor{gray!15} Shortened-Taylor      & \cellcolor{gray!15} 74.83 & \cellcolor{gray!15} 42.80 & \cellcolor{gray!15} 76.49 & \cellcolor{gray!15} \textbf{77.03} & \cellcolor{gray!15} \underline{77.62} & \cellcolor{gray!15} \underline{80.05} & \cellcolor{gray!15} \underline{53.92} & \cellcolor{gray!15} \underline{68.96} \\
        & \cellcolor{cyan!10} LLM-Pruner             & \cellcolor{cyan!10} 71.71 & \cellcolor{cyan!10} 38.40 & \cellcolor{cyan!10} 75.40 & \cellcolor{cyan!10} 62.98 & \cellcolor{cyan!10} 68.86 & \cellcolor{cyan!10} 74.11 & \cellcolor{cyan!10} 43.77 & \cellcolor{cyan!10} 62.18 \\
        & \cellcolor{cyan!10} SliceGPT               & \cellcolor{cyan!10} \textbf{82.29} & \cellcolor{cyan!10} 38.00 & \cellcolor{cyan!10} 76.82 & \cellcolor{cyan!10} 72.45 & \cellcolor{cyan!10} 72.63 & \cellcolor{cyan!10} 75.42 & \cellcolor{cyan!10} 49.19 & \cellcolor{cyan!10} 66.69 \\
        & \cellcolor{lime!15} MOD                   & \cellcolor{lime!15} 69.78 & \cellcolor{lime!15} 36.00 & \cellcolor{lime!15} 72.85 & \cellcolor{lime!15} 66.14 & \cellcolor{lime!15} 73.43 & \cellcolor{lime!15} 74.62 & \cellcolor{lime!15} 47.95 & \cellcolor{lime!15} 62.97 \\
        & \cellcolor{lime!15} D-LLM                 & \cellcolor{lime!15} 64.64 & \cellcolor{lime!15} 27.20 & \cellcolor{lime!15} 58.65 & \cellcolor{lime!15} 56.19 & \cellcolor{lime!15} 60.66 & \cellcolor{lime!15} 64.52 & \cellcolor{lime!15} 37.71 & \cellcolor{lime!15} 52.80 \\
        & \cellcolor{lime!15} SkipGPT               & \cellcolor{lime!15} \underline{80.39} & \cellcolor{lime!15} \textbf{47.20} & \cellcolor{lime!15} \underline{77.80} & \cellcolor{lime!15} 74.11 & \cellcolor{lime!15} \textbf{78.62} & \cellcolor{lime!15} \textbf{85.56} & \cellcolor{lime!15} \textbf{60.40} & \cellcolor{lime!15} \textbf{72.01} \\
        \cmidrule(lr){1-10}
        \multirow{9}{*}{40.0\%} 
        & \cellcolor{gray!15} SLEB              & \cellcolor{gray!15} 64.01 & \cellcolor{gray!15} 32.60 & \cellcolor{gray!15} 70.78 & \cellcolor{gray!15} 62.27 & \cellcolor{gray!15} 58.92 & \cellcolor{gray!15} 64.90 & \cellcolor{gray!15} 37.12 & \cellcolor{gray!15} 55.80 \\
        & \cellcolor{gray!15} ShortGPT              & \cellcolor{gray!15} 67.82 & \cellcolor{gray!15} 29.40 & \cellcolor{gray!15} 67.62 & \cellcolor{gray!15} 68.19 & \cellcolor{gray!15} 60.00 & \cellcolor{gray!15} 60.85 & \cellcolor{gray!15} 39.07 & \cellcolor{gray!15} 56.14 \\
        & \cellcolor{gray!15} Shortened-PPL         & \cellcolor{gray!15} 45.65 & \cellcolor{gray!15} \underline{33.80} & \cellcolor{gray!15} \underline{71.81} & \cellcolor{gray!15} 53.98 & \cellcolor{gray!15} 56.50 & \cellcolor{gray!15} 66.87 & \cellcolor{gray!15} 35.32 & \cellcolor{gray!15} 51.99 \\
        & \cellcolor{gray!15} Shortened-Taylor      & \cellcolor{gray!15} 73.60 & \cellcolor{gray!15} 30.80 & \cellcolor{gray!15} 69.04 & \cellcolor{gray!15} \underline{70.40} & \cellcolor{gray!15} 63.31 & \cellcolor{gray!15} 65.15 & \cellcolor{gray!15} 40.52 & \cellcolor{gray!15} \underline{58.97} \\
        & \cellcolor{cyan!10} LLM-Pruner             & \cellcolor{cyan!10} 63.24 & \cellcolor{cyan!10} 30.80 & \cellcolor{cyan!10} 66.53 & \cellcolor{cyan!10} 55.01 & \cellcolor{cyan!10} 48.05 & \cellcolor{cyan!10} 56.31 & \cellcolor{cyan!10} 30.80 & \cellcolor{cyan!10} 50.11 \\
        & \cellcolor{cyan!10} SliceGPT               & \cellcolor{cyan!10} \underline{74.77} & \cellcolor{cyan!10} 29.80 & \cellcolor{cyan!10} 63.65 & \cellcolor{cyan!10} 61.01 & \cellcolor{cyan!10} 51.02 & \cellcolor{cyan!10} 55.93 & \cellcolor{cyan!10} 33.87 & \cellcolor{cyan!10} 52.86 \\
        & \cellcolor{lime!15} MOD                   & \cellcolor{lime!15} 64.18 & \cellcolor{lime!15} 32.20 & \cellcolor{lime!15} 69.85 & \cellcolor{lime!15} 62.27 & \cellcolor{lime!15} \underline{65.67} & \cellcolor{lime!15} \underline{68.98} & \cellcolor{lime!15} \underline{42.49} & \cellcolor{lime!15} 57.95 \\
        & \cellcolor{lime!15} D-LLM                 & \cellcolor{lime!15} 58.13 & \cellcolor{lime!15} 26.60 & \cellcolor{lime!15} 52.72 & \cellcolor{lime!15} 54.14 & \cellcolor{lime!15} 41.58 & \cellcolor{lime!15} 46.96 & \cellcolor{lime!15} 28.92 & \cellcolor{lime!15} 44.15 \\
        & \cellcolor{lime!15} SkipGPT               & \cellcolor{lime!15} \textbf{81.74} & \cellcolor{lime!15} \textbf{41.20} & \cellcolor{lime!15} \textbf{77.31} & \cellcolor{lime!15} \textbf{75.29} & \cellcolor{lime!15} \textbf{82.01} & \cellcolor{lime!15} \textbf{86.44} & \cellcolor{lime!15} \textbf{60.58} & \cellcolor{lime!15} \textbf{72.08} \\
        \cmidrule(lr){1-10}
        \multirow{9}{*}{60.0\%} 
        & \cellcolor{gray!15} SLEB              & \cellcolor{gray!15} 60.15 & \cellcolor{gray!15} 27.00 & \cellcolor{gray!15} 62.35 & \cellcolor{gray!15} 54.22 & \cellcolor{gray!15} 38.89 & \cellcolor{gray!15} 47.22 & \cellcolor{gray!15} 26.54 & \cellcolor{gray!15} 45.20 \\
        & \cellcolor{gray!15} ShortGPT              & \cellcolor{gray!15} 60.27 & \cellcolor{gray!15} 26.60 & \cellcolor{gray!15} 57.23 & \cellcolor{gray!15} 52.40 & \cellcolor{gray!15} 35.56 & \cellcolor{gray!15} 35.56 & \cellcolor{gray!15} 23.12 & \cellcolor{gray!15} 41.53 \\
        & \cellcolor{gray!15} Shortened-PPL         & \cellcolor{gray!15} 60.42 & \cellcolor{gray!15} \underline{28.00} & \cellcolor{gray!15} \underline{63.11} & \cellcolor{gray!15} 51.53 & \cellcolor{gray!15} 32.81 & \cellcolor{gray!15} \underline{49.24} & \cellcolor{gray!15} 26.36 & \cellcolor{gray!15} 44.50 \\
        & \cellcolor{gray!15} Shortened-Taylor      & \cellcolor{gray!15} 55.41 & \cellcolor{gray!15} 27.80 & \cellcolor{gray!15} 60.33 & \cellcolor{gray!15} \underline{55.80} & \cellcolor{gray!15} 38.47 & \cellcolor{gray!15} 41.75 & \cellcolor{gray!15} 24.74 & \cellcolor{gray!15} 43.47 \\
        & \cellcolor{cyan!10} LLM-Pruner             & \cellcolor{cyan!10} 53.51 & \cellcolor{cyan!10} 26.20 & \cellcolor{cyan!10} 60.88 & \cellcolor{cyan!10} 51.38 & \cellcolor{cyan!10} 32.41 & \cellcolor{cyan!10} 41.49 & \cellcolor{cyan!10} 20.90 & \cellcolor{cyan!10} 40.97 \\
        & \cellcolor{cyan!10} SliceGPT               & \cellcolor{cyan!10} \underline{63.12} & \cellcolor{cyan!10} 26.40 & \cellcolor{cyan!10} 57.72 & \cellcolor{cyan!10} 52.56 & \cellcolor{cyan!10} 35.44 & \cellcolor{cyan!10} 39.23 & \cellcolor{cyan!10} 23.98 & \cellcolor{cyan!10} 42.64 \\
        & \cellcolor{lime!15} MOD                   & \cellcolor{lime!15} 59.69 & \cellcolor{lime!15} 27.80 & \cellcolor{lime!15} 55.60 & \cellcolor{lime!15} 54.06 & \cellcolor{lime!15} \underline{47.17} & \cellcolor{lime!15} 48.48 & \cellcolor{lime!15} \underline{31.14} & \cellcolor{lime!15} \underline{46.28} \\
        & \cellcolor{lime!15} D-LLM                 & \cellcolor{lime!15} 57.06 & \cellcolor{lime!15} 24.80 & \cellcolor{lime!15} 52.06 & \cellcolor{lime!15} 50.43 & \cellcolor{lime!15} 31.98 & \cellcolor{lime!15} 37.20 & \cellcolor{lime!15} 23.97 & \cellcolor{lime!15} 39.64 \\
        & \cellcolor{lime!15} SkipGPT               & \cellcolor{lime!15} \textbf{83.21} & \cellcolor{lime!15} \textbf{39.60} & \cellcolor{lime!15} \textbf{77.14} & \cellcolor{lime!15} \textbf{73.79} & \cellcolor{lime!15} \textbf{81.64} & \cellcolor{lime!15} \textbf{86.32} & \cellcolor{lime!15} \textbf{60.66} & \cellcolor{lime!15} \textbf{71.77} \\
        \specialrule{1.5pt}{0pt}{0pt}
    \end{tabular}

    \caption{Performance on classification tasks under different pruning ratio. 
        The dense baseline is Llama~3.1-Tulu-3-8B-SFT(LLM-insturct). 
        For each pruning ratio, the best result is marked in \textbf{bold}, and the second-best is \underline{underlined}. 
        Color coding indicates pruning strategy: 
        \colorbox{gray!15}{\rule[0.0ex]{0pt}{0.1em}Static Depth Pruning}, 
        \colorbox{cyan!10}{\rule[0.0ex]{0pt}{0.1em}Static Width Pruning}, 
        and \colorbox{lime!15}{\rule[0.0ex]{0pt}{0.1em}Dynamic Depth Pruning}.}
    \label{tab:table_classification}
    \vspace{-10pt}
\end{table*}

For evaluation, we assess LLM-instruct models on IFEval, HumanEval+, and classification tasks, and evaluate LLM-think models on GSM8K using CoT prompting. The results in \autoref{fig:calibration_and_post_trainng} consistently show that, for both LLM-instruct and LLM-think, using the original training corpus for calibration and post-pruning recovery yields the strongest performance. While general-domain and instruction-tuning datasets are sufficient to preserve basic linguistic competence, reflected by relatively stable performance on simpler benchmarks, they fail to retain more complex capabilities, resulting in near-zero performance on challenging tasks. In contrast, calibration and recovery using the original training corpus consistently preserve strong performance across all tasks.

\mybox{{\bf Conclusion}}{gray!40}{gray!10}{
    For both LLM-instruct and LLM-think, aligning pruning calibration and post-pruning recovery with the model’s original training corpus consistently yields superior preservation of pre-pruning capabilities.
}

\section{Effects of Pruning}
After confirming that using the original training corpus for both pruning calibration and post-pruning recovery more effectively recovers the performance of pruned models, we adopt this configuration as our default setting in the remainder of the paper. Specifically, we fix the calibration and recovery data to match the original training distribution, thereby controlling for dataset choice and isolating the effects of different pruning strategies in subsequent analyses. 

\begin{table*}[!ht]
    \scriptsize
    \setlength{\tabcolsep}{6pt}
    \renewcommand{\arraystretch}{1}
    \centering
    \begin{tabular}{c|l|ccccc|c|ccccc|c}
        \specialrule{1.5pt}{0pt}{0pt}
        \multirow{3}{*}{\textbf{Ratio}} 
        & \multirow{3}{*}{\textbf{Method}} 
        & \multicolumn{5}{c|}{\textbf{LLM-instruct: Generation (w/o CoT)}} & \multirow{3}{*}{\textbf{Avg.}\,$\uparrow$} 
        & \multicolumn{5}{c|}{\textbf{LLM-think: Reasoning Tasks}} & \multirow{3}{*}{\textbf{Avg.}\,$\uparrow$} \\
        \cmidrule(lr){3-7}\cmidrule(lr){9-13}
        & & IFE & TQA & PQA & HE & HE+ 
        & & MATH & AIME & LCB & GPQA & JEE & \\
        & & \textit{Pr.} & \textit{mc2} &  & \textit{p@10} & \textit{p@10}
        & & \textit{Acc} & \textit{Acc} & \textit{Acc} & \textit{Acc} & \textit{Acc} & \\
        \midrule
        0.00\% & Dense & 74.12 & 46.78 & 29.44 & 84.22 & 77.49 & 62.41 & 71.80 & 20.00 & 10.03 & 42.42 & 32.33 & 35.72 \\
        \cmidrule(lr){1-14}
        
        \multirow{9}{*}{20.0\%}
        & \cellcolor{gray!15} SLEB         & \cellcolor{gray!15} 59.51 & \cellcolor{gray!15} 41.68 & \cellcolor{gray!15} 15.71 & \cellcolor{gray!15} 67.79 & \cellcolor{gray!15} 60.89 & \cellcolor{gray!15} 49.12 & \cellcolor{gray!15} 49.00 & \cellcolor{gray!15} 3.33 & \cellcolor{gray!15} 1.08 & \cellcolor{gray!15} 31.81 & \cellcolor{gray!15} 14.22 & \cellcolor{gray!15} 19.89 \\
        & \cellcolor{gray!15} ShortGPT         & \cellcolor{gray!15} 60.81 & \cellcolor{gray!15} 45.24 & \cellcolor{gray!15} 13.16 & \cellcolor{gray!15} 66.82 & \cellcolor{gray!15} 60.88 & \cellcolor{gray!15} 49.38 & \cellcolor{gray!15} 52.00 & \cellcolor{gray!15} 3.33 & \cellcolor{gray!15} \textbf{3.25} & \cellcolor{gray!15} 23.74 & \cellcolor{gray!15} \textbf{19.37} & \cellcolor{gray!15} \underline{20.34} \\
        & \cellcolor{gray!15} Shortened-PPL    & \cellcolor{gray!15} 48.42 & \cellcolor{gray!15} 36.00 & \cellcolor{gray!15} \textbf{21.55} & \cellcolor{gray!15} 42.77 & \cellcolor{gray!15} 40.33 & \cellcolor{gray!15} 37.81 & \cellcolor{gray!15} 53.00 & \cellcolor{gray!15} 0.00 & \cellcolor{gray!15} 1.63 & \cellcolor{gray!15} \textbf{29.80} & \cellcolor{gray!15} 17.86 & \cellcolor{gray!15} \textbf{20.46} \\
        & \cellcolor{gray!15} Shortened-Taylor & \cellcolor{gray!15} 65.61 & \cellcolor{gray!15} \underline{45.47} & \cellcolor{gray!15} 15.31 & \cellcolor{gray!15} 80.63 & \cellcolor{gray!15} 73.36 & \cellcolor{gray!15} 56.08 & \cellcolor{gray!15} \underline{53.40} & \cellcolor{gray!15} 3.33 & \cellcolor{gray!15} 0.00 & \cellcolor{gray!15} 23.74 & \cellcolor{gray!15} 19.08 & \cellcolor{gray!15} 19.91 \\
        & \cellcolor{cyan!10} LLM-Pruner          & \cellcolor{cyan!10} 52.49 & \cellcolor{cyan!10} 42.62 & \cellcolor{cyan!10} 15.16 & \cellcolor{cyan!10} 57.29 & \cellcolor{cyan!10} 53.34 & \cellcolor{cyan!10} 44.18 & \cellcolor{cyan!10} 52.30 & \cellcolor{cyan!10} 3.33 & \cellcolor{cyan!10} \underline{1.85} & \cellcolor{cyan!10} 22.50 & \cellcolor{cyan!10} 5.35  &  \cellcolor{cyan!10} 17.07 \\
        & \cellcolor{cyan!10} SliceGPT         & \cellcolor{cyan!10} 63.58 & \cellcolor{cyan!10} 44.74 & \cellcolor{cyan!10} 10.77 & \cellcolor{cyan!10} 74.71 & \cellcolor{cyan!10} 70.47 & \cellcolor{cyan!10} 52.85 & \cellcolor{cyan!10} \textbf{60.00} & \cellcolor{cyan!10} 3.33 & \cellcolor{cyan!10} 0.00 & \cellcolor{cyan!10} 22.50 & \cellcolor{cyan!10} 6.69  & \cellcolor{cyan!10} 18.50 \\
        & \cellcolor{lime!15} MOD                & \cellcolor{lime!15} \textbf{68.20} & \cellcolor{lime!15} 44.74 & \cellcolor{lime!15} 14.75 & \cellcolor{lime!15} \underline{82.18} & \cellcolor{lime!15} \textbf{78.40} & \cellcolor{lime!15} \underline{57.65} & \cellcolor{lime!15} 38.40 & \cellcolor{lime!15} 0.00 & \cellcolor{lime!15} 0.00 & \cellcolor{lime!15} 22.50  & \cellcolor{lime!15} 10.71 & \cellcolor{lime!15} 14.32 \\
        & \cellcolor{lime!15} D-LLM            & \cellcolor{lime!15} 29.75 & \cellcolor{lime!15} 42.63 & \cellcolor{lime!15} 10.94 & \cellcolor{lime!15} 26.68 & \cellcolor{lime!15} 20.47 & \cellcolor{lime!15} 26.09 & \cellcolor{lime!15} 1.00  & \cellcolor{lime!15} 0.00 & \cellcolor{lime!15} 0.00  & \cellcolor{lime!15} 15.00 &  \cellcolor{lime!15} 4.17 &  \cellcolor{lime!15} 4.03\\
        & \cellcolor{lime!15} SkipGPT          & \cellcolor{lime!15} \underline{67.09} & \cellcolor{lime!15} \textbf{47.03} & \cellcolor{lime!15} \underline{17.10} & \cellcolor{lime!15} \textbf{83.65} & \cellcolor{lime!15} \underline{76.12} & \cellcolor{lime!15} \textbf{58.20} & \cellcolor{lime!15} 0.00 & \cellcolor{lime!15} 0.00 & \cellcolor{lime!15} 0.00 & \cellcolor{lime!15} 13.63 & \cellcolor{lime!15} 0.00 &  \cellcolor{lime!15} 2.73 \\
        \cmidrule(lr){1-14}
        
        \multirow{9}{*}{40.0\%}
        & \cellcolor{gray!15} SLEB         & \cellcolor{gray!15} 36.78 & \cellcolor{gray!15} 40.70 & \cellcolor{gray!15} 10.30  & \cellcolor{gray!15} 25.43 & \cellcolor{gray!15} 23.05 & \cellcolor{gray!15} 27.25 & \cellcolor{gray!15} 4.40 & \cellcolor{gray!15} 0.00 & \cellcolor{gray!15} 0.00 & \cellcolor{gray!15} \textbf{29.29} & \cellcolor{gray!15} \underline{8.59} & \cellcolor{gray!15} 8.46 \\
        & \cellcolor{gray!15} ShortGPT         & \cellcolor{gray!15} 31.60 & \cellcolor{gray!15} 46.89 & \cellcolor{gray!15} 9.44  & \cellcolor{gray!15} 13.71 & \cellcolor{gray!15} 11.88 & \cellcolor{gray!15} 22.70 & \cellcolor{gray!15} 3.40 & \cellcolor{gray!15} 0.00 & \cellcolor{gray!15} 0.00 & \cellcolor{gray!15} 15.15 & \cellcolor{gray!15} \underline{4.27} & \cellcolor{gray!15} 4.56 \\
        & \cellcolor{gray!15} Shortened-PPL    & \cellcolor{gray!15} 27.91 & \cellcolor{gray!15} 37.34 & \cellcolor{gray!15} 13.49 & \cellcolor{gray!15} 18.92 & \cellcolor{gray!15} 15.15 & \cellcolor{gray!15} 22.56 & \cellcolor{gray!15} 0.40 & \cellcolor{gray!15} 0.00 & \cellcolor{gray!15} 0.00 & \cellcolor{gray!15} 13.64 & \cellcolor{gray!15} 2.52 & \cellcolor{gray!15} 3.31 \\
        & \cellcolor{gray!15} Shortened-Taylor & \cellcolor{gray!15} 50.09 & \cellcolor{gray!15} 43.69 & \cellcolor{gray!15} 10.92 & \cellcolor{gray!15} 46.18 & \cellcolor{gray!15} 41.11 & \cellcolor{gray!15} 38.40 & \cellcolor{gray!15} 3.80 & \cellcolor{gray!15} 0.00 & \cellcolor{gray!15} 0.00 & \cellcolor{gray!15} 16.16 & \cellcolor{gray!15} 2.86 & \cellcolor{gray!15} 4.56 \\
        & \cellcolor{cyan!10} LLM-Pruner          & \cellcolor{cyan!10} 36.41 & \cellcolor{cyan!10} 42.98 & \cellcolor{cyan!10} 9.60  & \cellcolor{cyan!10} 23.90 & \cellcolor{cyan!10} 22.19 & \cellcolor{cyan!10} 27.02 & \cellcolor{cyan!10} 0.00 & \cellcolor{cyan!10} \textbf{3.33} & \cellcolor{cyan!10} 0.00 & \cellcolor{cyan!10} 20.00 & \cellcolor{cyan!10} 0.00 & \cellcolor{cyan!10} 4.67 \\
        & \cellcolor{cyan!10} SliceGPT         & \cellcolor{cyan!10} \underline{57.30}& \cellcolor{cyan!10} \underline{48.12} & \cellcolor{cyan!10} 6.83  & \cellcolor{cyan!10} 60.59 & \cellcolor{cyan!10} 55.10 & \cellcolor{cyan!10} 45.59 & \cellcolor{cyan!10} \underline{29.00}  & \cellcolor{cyan!10} 0.00 & \cellcolor{cyan!10} 0.00 & \cellcolor{cyan!10} \underline{25.00} & \cellcolor{cyan!10} 2.37 & \cellcolor{cyan!10} \textbf{11.27}  \\
        & \cellcolor{lime!15} MOD              & \cellcolor{lime!15} 48.79 & \cellcolor{lime!15} 42.26 & \cellcolor{lime!15} \underline{14.08} & \cellcolor{lime!15} \underline{72.23} & \cellcolor{lime!15} \underline{67.77} & \cellcolor{lime!15} \underline{49.03} & \cellcolor{lime!15} \textbf{33.66} & \cellcolor{lime!15} 0.00 & \cellcolor{lime!15} 0.00 & \cellcolor{lime!15} 13.75 & \cellcolor{lime!15} \textbf{6.69} &  \cellcolor{lime!15} \underline{10.82} \\
        & \cellcolor{lime!15} D-LLM            & \cellcolor{lime!15} 20.14 & \cellcolor{lime!15} 44.39 & \cellcolor{lime!15} 7.63  & \cellcolor{lime!15} 11.63 & \cellcolor{lime!15} 8.67  & \cellcolor{lime!15} 18.49 & \cellcolor{lime!15} 3.00  & \cellcolor{lime!15} 0.00 & \cellcolor{lime!15} 0.00  & \cellcolor{lime!15} 17.50 &  \cellcolor{lime!15} 3.78 & \cellcolor{lime!15} 4.86 \\
        & \cellcolor{lime!15} SkipGPT            & \cellcolor{lime!15} \textbf{70.61} & \cellcolor{lime!15} \textbf{50.04} & \cellcolor{lime!15} \textbf{23.88} & \cellcolor{lime!15} \textbf{83.09} & \cellcolor{lime!15} \textbf{75.67} & \cellcolor{lime!15} \textbf{60.66}  & \cellcolor{lime!15} 0.00 &  \cellcolor{lime!15} 0.00 &  \cellcolor{lime!15} 0.00  & \cellcolor{lime!15} 10.33 & \cellcolor{lime!15} 0.00  & \cellcolor{lime!15} 2.07  \\
        \cmidrule(lr){1-14}
        
        \multirow{9}{*}{60.0\%}
        & \cellcolor{gray!15} SLEB         & \cellcolor{gray!15} 15.15 & \cellcolor{gray!15} 44.46 & \cellcolor{gray!15} 7.37  & \cellcolor{gray!15} 7.39  & \cellcolor{gray!15} 5.37  & \cellcolor{gray!15} 15.95 & \cellcolor{gray!15} 0.00 & \cellcolor{gray!15} 0.00 & \cellcolor{gray!15} 0.00 & \cellcolor{gray!15} 11.36  & \cellcolor{gray!15} 0.00 & \cellcolor{gray!15} 2.27  \\
        & \cellcolor{gray!15} ShortGPT         & \cellcolor{gray!15} 10.90 & \cellcolor{gray!15} 47.10 & \cellcolor{gray!15} 5.69  & \cellcolor{gray!15} 0.00  & \cellcolor{gray!15} 0.30  & \cellcolor{gray!15} 12.80 & \cellcolor{gray!15} 0.00 & \cellcolor{gray!15} 0.00 & \cellcolor{gray!15} 0.00 & \cellcolor{gray!15} 1.51  & \cellcolor{gray!15} 0.00 & \cellcolor{gray!15} 0.30  \\
        & \cellcolor{gray!15} Shortened-PPL    & \cellcolor{gray!15} 18.11 & \cellcolor{gray!15} 41.99 & \cellcolor{gray!15} 5.40  & \cellcolor{gray!15} 7.37  & \cellcolor{gray!15} 6.17  & \cellcolor{gray!15} 15.81 & \cellcolor{gray!15} 0.00 & \cellcolor{gray!15} 0.00 & \cellcolor{gray!15} 0.00 & \cellcolor{gray!15} 12.46 & \cellcolor{gray!15} 0.00 & \cellcolor{gray!15} 2.49  \\
        & \cellcolor{gray!15} Shortened-Taylor & \cellcolor{gray!15} 17.56 & \cellcolor{gray!15} 43.74 & \cellcolor{gray!15} 6.46  & \cellcolor{gray!15} 6.99  & \cellcolor{gray!15} 5.13  & \cellcolor{gray!15} 15.98 & \cellcolor{gray!15} 0.00 & \cellcolor{gray!15} 0.00 & \cellcolor{gray!15} 0.00 & \cellcolor{gray!15} \textbf{16.66} & \cellcolor{gray!15} 0.00 & \cellcolor{gray!15} 3.33  \\
        & \cellcolor{cyan!10} LLM-Pruner       & \cellcolor{cyan!10} 20.14 & \cellcolor{cyan!10} 47.19 & \cellcolor{cyan!10} 2.29  & \cellcolor{cyan!10} 6.58  & \cellcolor{cyan!10} 6.39  & \cellcolor{cyan!10} 16.52 & \cellcolor{cyan!10} 0.00 & \cellcolor{cyan!10} 0.00 & \cellcolor{cyan!10} 0.00 & \cellcolor{cyan!10} \underline{13.75} &  \cellcolor{cyan!10} 0.00 & \cellcolor{cyan!10} 2.75  \\
        & \cellcolor{cyan!10} SliceGPT         & \cellcolor{cyan!10} \underline{35.12} & \cellcolor{cyan!10} \textbf{47.54} & \cellcolor{cyan!10} \underline{6.61} & \cellcolor{cyan!10} 29.00 & \cellcolor{cyan!10} 25.59 & \cellcolor{cyan!10} 28.77 & \cellcolor{cyan!10} \underline{3.00}  & \cellcolor{cyan!10} 0.00 & \cellcolor{cyan!10} 0.00 & \cellcolor{cyan!10} 2.00 & \cellcolor{cyan!10} 0.67 & \cellcolor{cyan!10} 1.13 \\
        & \cellcolor{lime!15} MOD                & \cellcolor{lime!15} 10.35 & \cellcolor{lime!15} 44.16 & \cellcolor{lime!15} 3.73  & \cellcolor{lime!15} \underline{55.93} & \cellcolor{lime!15} \underline{48.39} & \cellcolor{lime!15} \underline{32.51} & \cellcolor{lime!15} \textbf{7.00} &\cellcolor{lime!15} 0.00 & \cellcolor{lime!15} 0.00 & \cellcolor{lime!15} \underline{13.75} & \cellcolor{lime!15} 2.23 & \cellcolor{lime!15} \textbf{4.60} \\
        & \cellcolor{lime!15} D-LLM            & \cellcolor{lime!15} 11.46 & \cellcolor{lime!15} \underline{47.22} & \cellcolor{lime!15} 3.74  & \cellcolor{lime!15} 2.31  & \cellcolor{lime!15} 2.10  & \cellcolor{lime!15} 13.37 & \cellcolor{lime!15} \underline{3.00}  & \cellcolor{lime!15} 0.00 & \cellcolor{lime!15} 0.00  & \cellcolor{lime!15} 11.25 &  \cellcolor{lime!15} 5.38 &  \cellcolor{lime!15} \underline{3.93} \\
        & \cellcolor{lime!15} SkipGPT            & \cellcolor{lime!15} \textbf{69.68} & \cellcolor{lime!15} 45.19 & \cellcolor{lime!15} \textbf{22.97} & \cellcolor{lime!15} \textbf{83.15} & \cellcolor{lime!15} \textbf{77.25} & \cellcolor{lime!15} \textbf{59.65} &   \cellcolor{lime!15} 0.00 &  \cellcolor{lime!15} 0.00 &  \cellcolor{lime!15} 0.00  & \cellcolor{lime!15} 2.83 & \cellcolor{lime!15} 0.00  & \cellcolor{lime!15} 0.57  \\
        \specialrule{1.5pt}{0pt}{0pt}
    \end{tabular}
    \caption{Performance on generation (w/o CoT) and reasoning tasks under different pruning ratio. 
    For each pruning ratio, the best result is in \textbf{bold}, and the second-best is \underline{underlined}.}
    \label{tab:final_results}
    \vspace{-10pt}
\end{table*}

\paragraph{Within static depth pruning methods,} we study four representative strategies: \textit{SLEB}, \textit{ShortGPT}, \textit{Shortened-PPL}, and \textit{Shortened-Taylor}. as shown in \autoref{tab:table_classification} and \autoref{tab:performance_decline}, for LLM-instruct, \textit{Shortened-Taylor} consistently outperforms competing methods under both 20\% and 40\% pruning ratios. This advantage is particularly pronounced on generation tasks, where the average performance decline remains below 7.04\% at a 20\% pruning ratio. 
Shortened-Taylor is particularly effective at identifying redundant layers for LLM-instruct. Its advantage comes from using a second-order Taylor approximation to estimate each layer’s contribution to the training loss, yielding a more sensitivity-aware notion of importance than purely local heuristics. By comparison, ShortGPT relies on input–output representation similarity, while Shortened-PPL ranks layers by the perplexity change induced by removing each layer. SLEB further introduces an iterative procedure that recomputes layer importance after every pruning step.
For LLM-think, however, a markedly different regime emerges. As shown in \autoref{tab:performance_decline}, the performance decline of all methods exceeds 42\% even at a 20\% pruning ratio, indicating that removing only a small number of layers (e.g., six) leads to substantial degradation. \textit{These results suggest that static depth pruning may not be an ideal pruning strategy for LLM-think.}

\begin{figure}[ht]
    \centering
    \includegraphics[width=1\linewidth]{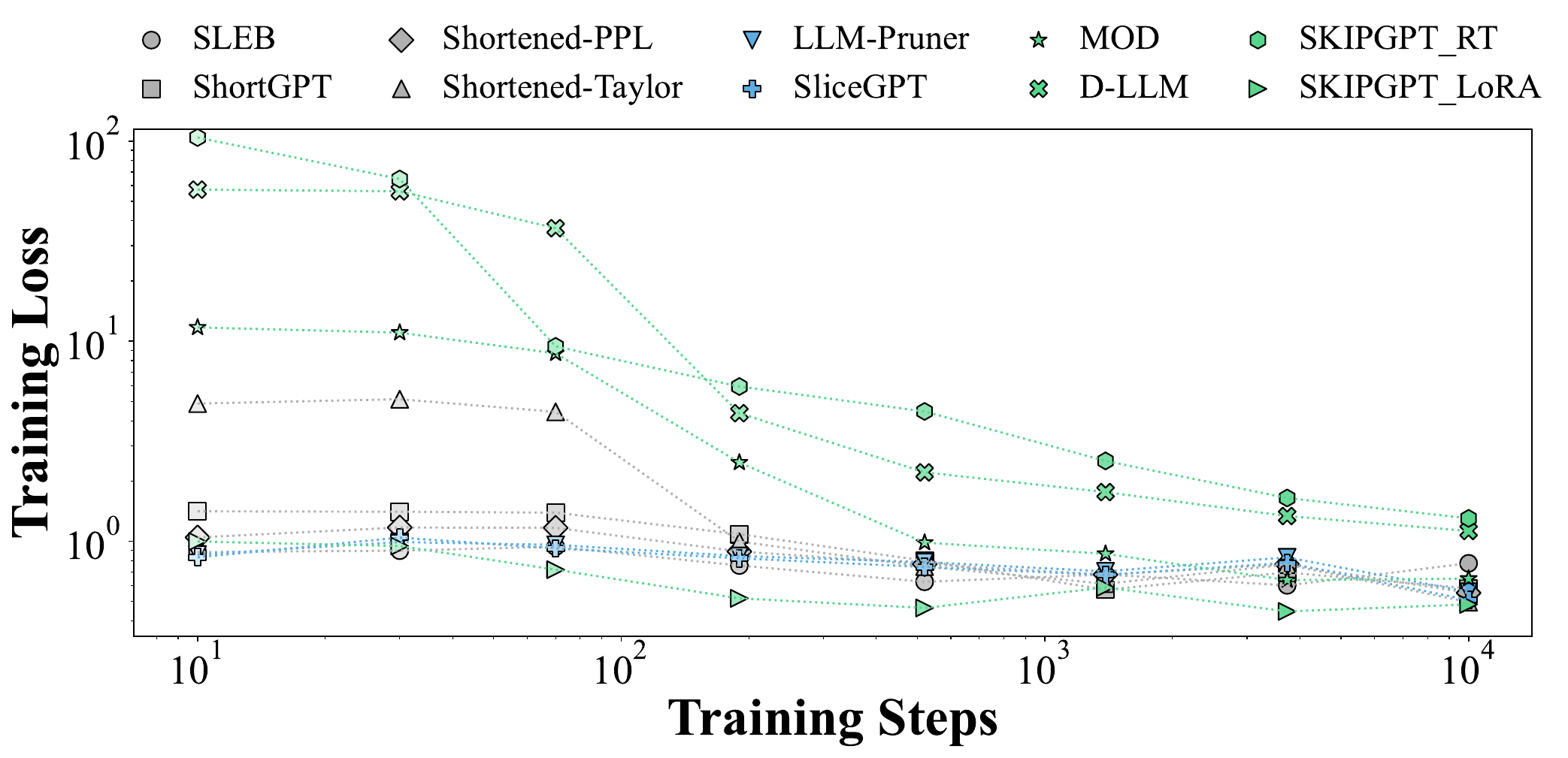}
    \caption{Training loss of \textbf{LLM-instruct} at 20\% pruning ratio. Dynamic methods exhibit substantially lower loss (with D-LLM as an exception).}
    \label{fig:train-loss_0.2-instruct}
\end{figure}

\begin{figure*}[ht] 
    \centering
    \includegraphics[width=1\textwidth]{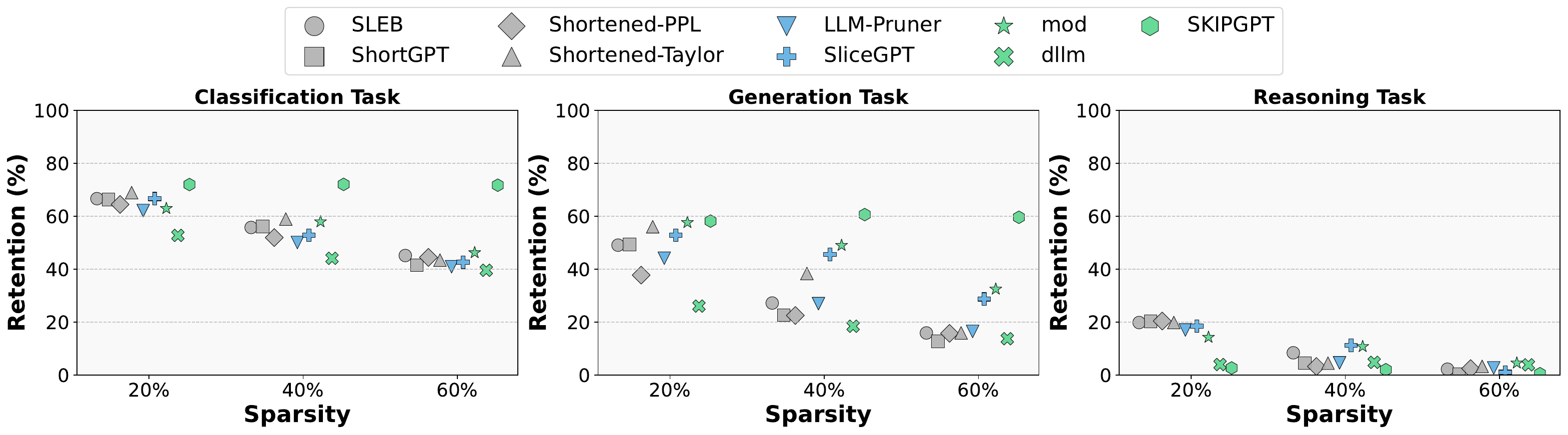} 
    \vspace{-20pt}
    \caption{Performance of different pruning methods under varying pruning ratios on classification, generation, and reasoning tasks.}
    \label{fig:retention}
\end{figure*}

\paragraph{Within static width pruning methods,} we compare two representative strategies: \textit{LLM-Pruner} and \textit{SliceGPT}. 
As shown in \autoref{fig:retention}, SliceGPT consistently outperforms LLM-Pruner across all pruning ratios on both LLM-instruct and LLM-think. \autoref{tab:performance_decline} shows that, on classification tasks, SliceGPT maintains a modest yet stable advantage, with the performance gap remaining small and largely insensitive to higher pruning ratios. In contrast, on generation and reasoning tasks, the gap widens substantially as pruning becomes more aggressive.
One plausible explanation is the difference in their width-reduction mechanisms. SliceGPT performs subspace-based compression by estimating a low-dimensional subspace from calibration activations and removing low-variance directions, thereby shrinking hidden dimensions in a globally consistent manner. In contrast, LLM-Pruner removes structured modules (e.g., attention- or MLP-related groups) based on importance scores, potentially discarding discrete functional components that are harder to preserve under high sparsity. \textit{Overall, these results suggest that subspace-based width reduction (SliceGPT) preserves model performance more reliably than structured component deletion (LLM-Pruner) across both LLM-instruct and LLM-think.}

\begin{table*}[!t]
    \centering
    \resizebox{\textwidth}{!}{
        \begin{tabular}{l*{3}{l}|*{3}{l}|*{3}{l}}
            \toprule
            & \multicolumn{3}{c}{\tf{Classification}} 
            & \multicolumn{3}{c}{\tf{Generation (w/o CoT)}} 
            & \multicolumn{3}{c}{\tf{Reasoning}} \\
            \cmidrule(lr){2-4} \cmidrule(lr){5-7} \cmidrule(lr){8-10}
            \rowcolor{gray!10}
            Dense 
            & \multicolumn{3}{c|}{74.18} 
            & \multicolumn{3}{c|}{62.41} 
            & \multicolumn{3}{c}{35.72} \\
            \midrule
            \multicolumn{1}{r}{Sparsity} 
            & \multicolumn{1}{c}{0.2} & \multicolumn{1}{c}{0.4} & \multicolumn{1}{c|}{0.6}
            & \multicolumn{1}{c}{0.2} & \multicolumn{1}{c}{0.4} & \multicolumn{1}{c|}{0.6}
            & \multicolumn{1}{c}{0.2} & \multicolumn{1}{c}{0.4} & \multicolumn{1}{c}{0.6} \\
            \midrule
            SLEB       & 66.68\da{10.11} & 55.80\da{24.78} & 45.20\da{39.07} 
            & 49.12\da{21.29} & 27.25\da{56.34} & 15.95\da{74.44}
            & 19.89\da{44.32} & 8.46\da{76.32} & 2.27\da{93.65} \\   
            ShortGPT   & 66.35\da{10.56} & 56.14\da{24.32} & 41.53\da{44.01} 
            & 49.38\da{20.88} & 22.70\da{63.63} & 12.80\da{79.49}
            & 20.34\da{43.06} & 4.56\da{87.23} & 0.30\da{99.16} \\     
            Shortened-PPL & 64.49\da{13.06} & 51.99\da{29.91} & 44.50\da{40.01} 
            & 37.81\da{39.42} & 22.56\da{63.85} & 15.81\da{74.67} 
            & 20.46\da{42.72} & 3.31\da{90.73} & 2.49\da{93.03} \\    
            Shortened-Taylor & 68.96\da{7.04} & 58.97\da{20.50} & 43.47\da{41.40}   
            & 56.08\da{10.14} & 38.40\da{38.47} & 15.98\da{74.40} 
            & 19.91\da{44.26} & 4.56\da{87.23} & 3.33\da{90.68} \\ 
            \cmidrule(lr){2-10}
            \rowcolor{gray!15}
            \textbf{AD (avg)} 
            & \multicolumn{1}{c}{10.19\%} & \multicolumn{1}{c}{24.88\%} & \multicolumn{1}{c|}{41.12\%} 
            & \multicolumn{1}{c}{22.93\%} & \multicolumn{1}{c}{55.57\%} & \multicolumn{1}{c|}{75.75\%} 
            & \multicolumn{1}{c}{43.59\%} & \multicolumn{1}{c}{85.38\%} & \multicolumn{1}{c}{94.13\%} \\
            \midrule
            LLM-Pruner & 62.18\da{16.18} & 50.11\da{32.45} & 40.97\da{44.77} 
            & 44.18\da{29.21} & 27.02\da{56.71} & 16.52\da{73.53} 
            & 17.07\da{52.21} & \phantom{1}4.67\da{86.93} & 2.75\da{92.30} \\
            SliceGPT   & 66.69\da{10.10} & 52.86\da{28.74} & 42.64\da{42.52} 
            & 52.85\da{15.32} & 45.59\da{26.95} & 28.77\da{53.90}
            & 18.50\da{48.21} & 11.27\da{68.45} & 1.13\da{96.84} \\
            \cmidrule(lr){2-10}
            \rowcolor{gray!15}
            \textbf{AD (avg)} 
            & \multicolumn{1}{c}{13.14\%} & \multicolumn{1}{c}{30.60\%} & \multicolumn{1}{c|}{43.65\%}
            & \multicolumn{1}{c}{22.27\%} & \multicolumn{1}{c}{41.83\%} & \multicolumn{1}{c|}{63.72\%} 
            & \multicolumn{1}{c}{50.21\%} & \multicolumn{1}{c}{77.69\%} & \multicolumn{1}{c}{94.57\%} \\
            \midrule
            MOD     & 62.97\da{15.11} & 57.95\da{21.88} & 46.28\da{37.61} 
            & 57.65\da{7.63} & 49.03\da{21.44} & 32.51\da{47.91}
            & 14.32\da{59.91} & 10.82\da{69.71} & 4.60\da{87.12} \\
            D-LLM   & 52.80\da{28.82} & 44.15\da{40.48} & 39.64\da{46.56}
            & 26.09\da{58.20} & 18.49\da{70.37} & 13.77\da{77.94} 
            & \phantom{1}4.03\da{88.72} & \phantom{1}4.86\da{86.39} & 3.93\da{89.00} \\
            SkipGPT & 72.01\da{2.93} & 72.08\da{2.83} & 71.77\da{3.25} 
            & 58.20\da{6.75} & 60.66\da{2.80} & 59.65\da{4.42} 
            & \phantom{1}2.73\da{92.36} & \phantom{1}2.07\da{94.20} & 0.57\da{98.40} \\
            \cmidrule(lr){2-10}
            \rowcolor{gray!15}
            \textbf{AD (avg)} 
            & \multicolumn{1}{c}{15.62\%} & \multicolumn{1}{c}{21.73\%} & \multicolumn{1}{c|}{29.14\%} 
            & \multicolumn{1}{c}{24.19\%} & \multicolumn{1}{c}{31.54\%} & \multicolumn{1}{c|}{43.42\%}
            & \multicolumn{1}{c}{80.33\%} & \multicolumn{1}{c}{83.43\%} & \multicolumn{1}{c}{91.51\%} \\
            \bottomrule
        \end{tabular}
    }
    \caption{Performance decline of pruning methods on classification, generation, and reasoning tasks. The row Dense shows the unpruned model performance. Rows marked \textbf{average decline (AD)} shows the average relative performance drop of methods within the same pruning strategy at each sparsity level.}
    \label{tab:performance_decline}
\end{table*}

\begin{figure}[!ht]
    \centering
    \includegraphics[width=1\linewidth]{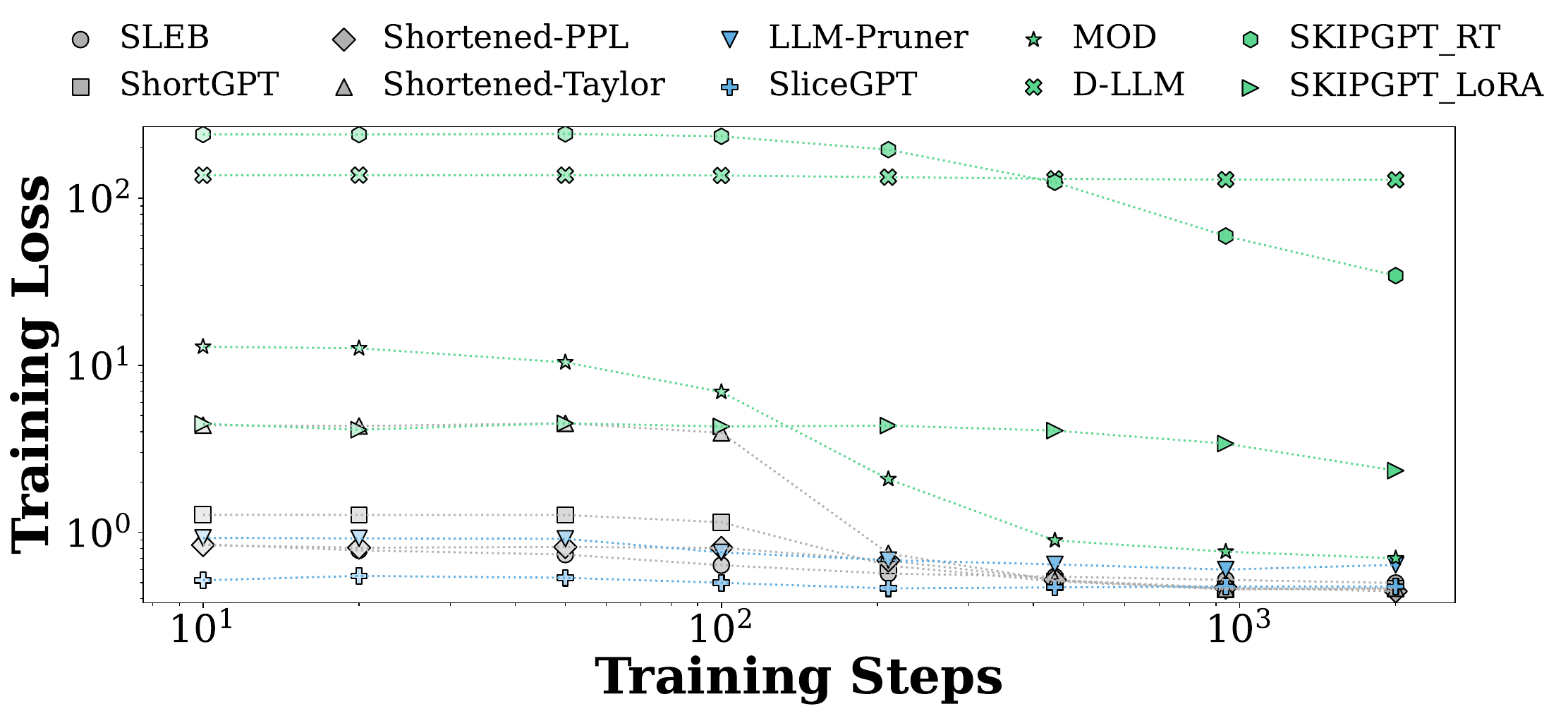}
    \caption{Training loss of \textbf{LLM-think} at 20\% pruning ratio. Dynamic methods exhibit substantially higher loss (with MOD as an exception), whereas static depth and width pruning remain lower and more stable.}
    \label{fig:train-loss_0.2}
\end{figure}

\paragraph{Within dynamic depth pruning methods,} we compare three representative strategies: MOD, D-LLM, and SkipGPT. As shown in \autoref{tab:performance_decline}, for LLM-instruct, SkipGPT consistently and substantially outperforms both D-LLM and MOD in performance retention across all pruning ratios. This advantage becomes more pronounced on generation tasks under higher pruning ratios; even at a 60\% pruning ratio, SkipGPT maintains an average performance decline below 4\% across both task types, achieving near-lossless compression.
For LLM-think, however, a markedly different trend emerges. Even at a 20\% pruning ratio, D-LLM and SkipGPT suffer severe semantic degradation, producing repetitive and incoherent outputs, whereas only MOD retains approximately 40\% of the dense model’s performance. \textit{This discrepancy raises an important question: why does SkipGPT, which achieves near-lossless performance on LLM-instruct, fail to generalize to LLM-think, while MOD preserves a non-trivial fraction of the model’s reasoning capability?}
As shown in \autoref{fig:train-loss_0.2-instruct} and \autoref{fig:train-loss_0.2}, SkipGPT exhibits contrasting optimization behavior across settings. On LLM-instruct, its LoRA fine-tuning stage converges to the lowest training loss among dynamic methods. On LLM-think, however, SkipGPT’s router-tuning stage converges to a substantially higher loss than MOD.
We hypothesize that this behavior may arise from SkipGPT’s highly flexible compute allocation: frequent token-wise routing changes can destabilize depth-wise computation trajectories, reducing path continuity across layers. In contrast, MOD enforces a Top-$k$ constrained routing policy that trades adaptivity for more consistent layer-wise structure. This fixed compute budget yields more stable optimization under predictable resource constraints, allowing MOD to preserve a substantial fraction of the model’s reasoning capability.
\textit{Overall, these results suggest that SkipGPT’s highly flexible token-wise routing works well for LLM-instruct but is less reliable for LLM-think, whereas MOD’s Top-$k$ constraint trades adaptivity for more stable reasoning retention.}

\paragraph{Static depth vs. width pruning.} For LLM-instruct, as shown in \autoref{tab:performance_decline} and \autoref{fig:retention}, static depth pruning slightly outperforms width pruning on classification tasks, whereas width pruning demonstrates greater robustness on generation tasks. While both approaches perform similarly at a 20\% pruning ratio, increasing the ratio to 40\% leads to a markedly larger performance drop for depth pruning (55.57\%) than for width pruning (41.83\%).
This pattern extends to LLM-think: despite marginally stronger performance from depth pruning at 20\%, both methods deteriorate rapidly at higher pruning levels. At a 40\% pruning ratio, depth pruning drops by 85.38\%, compared to 77.69\% for width pruning.
This gap reflects a fundamental structural difference between depth and width pruning. While depth pruning reduces the model’s depth, width pruning primarily limits representational capacity. As a result, width pruning exhibits more stable performance on generation and reasoning tasks at higher sparsity levels.
Overall, these results suggest that preserving model depth is more critical than preserving width, implying that depth should be pruned conservatively while width admits more aggressive compression.
\mybox{{\bf Conclusion}}{gray!40}{gray!10}{At higher sparsity levels, preserving model depth is more critical than preserving width.}

\paragraph{Static vs. Dynamic Pruning.}
As shown in \autoref{fig:retention}, on LLM-instruct, dynamic pruning methods consistently outperform static ones. In particular, both SkipGPT and MOD achieve superior performance across pruning ratios, with the advantage becoming more pronounced at higher pruning ratios.
In contrast, this trend reverses on LLM-think: dynamic pruning methods consistently underperform static pruning methods. As shown in \autoref{fig:train-loss_0.2}, under a 20\% pruning ratio, static pruning methods reliably converge to low-loss regions, whereas many dynamic methods display unstable or even pathological optimization dynamics and settle at substantially higher training loss. This optimization gap provides a mechanistic explanation for the weaker downstream performance of dynamic pruning on LLM-think. Notably, MOD is a clear exception among dynamic approaches, exhibiting strong and stable convergence that is comparable to static pruning methods.
These results indicate that reasoning performance hinges less on whether pruning is static or dynamic than on whether the pruning strategy preserves continuous computation paths along the depth dimension. Accordingly, static width pruning methods (e.g., LLM-Pruner, SliceGPT) and dynamic pruning(MOD) perform better on LLM-think.

\mybox{{\bf Conclusion}}{gray!40}{gray!10}{For LLM-instruct, dynamic pruning yields the best performance, whereas for LLM-think, static pruning retains higher performance.}

\section{Conclusion}
We study whether pruning methods developed for LLM-instruct generalize to LLM-think, where models explicitly generate long intermediate reasoning traces. Using a controlled pipeline that aligns calibration and post-pruning recovery with each model’s original training distribution, we construct matched model–dataset pairs for both paradigms and evaluate static depth pruning, static width pruning, and dynamic pruning across 17 classification, generation, and reasoning tasks. The results show clear paradigm and task-dependent trade-offs: static depth pruning is more effective for classification, whereas static width pruning degrades more gracefully for generation and reasoning. Dynamic pruning benefits classification and generation but is substantially harder to optimize for long-chain reasoning, making aggressive compute skipping risky for reasoning-intensive workloads. Overall, dynamic methods lead on classification and generation, while static pruning better preserves reasoning performance, highlighting an important gap in current dynamic pruning for LLM-think.

\section*{Limitations}
Our experiments are conducted on Llama-8B–scale language models, and we do not explicitly evaluate how the observed pruning behaviors generalize across model scales. While our findings reveal clear and consistent trends within this setting, extending the analysis to smaller and larger models remains an important direction for future work.

\bibliography{custom}

\clearpage
\appendix

\section{The Use of Large Language Models}
Large Language Models (LLMs) were used to support the writing and editing of this manuscript. Specifically, we employed an LLM to refine the language, improve readability, and enhance clarity in selected sections. The model was used for tasks such as sentence rephrasing, grammar correction, and improving the overall flow of the text.

The LLM was not involved in generating the study’s ideas, designing the research methodology, conducting experiments, analyzing data, or interpreting results. All scientific concepts, methods, and analyses presented in this work were independently conceived and carried out by the authors.

The authors take full responsibility for the content of the manuscript, including portions refined with the assistance of the LLM. Its use followed ethical standards and did not contribute to plagiarism or scientific misconduct.

\section{Related Work}
\label{sec:related_work}

The multi-layer Transformer architecture of LLMs inherently contains substantial parameter redundancy. Model pruning is a key technique to mitigate this redundancy by removing non-essential components. In practice, hardware-friendly structured pruning, which eliminates entire components such as layers, channels, or attention heads, has become the dominant approach and can be broadly categorized into static depth, static width, and dynamic depth methods.

\subsection{Static Depth Pruning}
Static depth pruning reduces model depth by removing fixed Transformer layers or their internal submodules (Attention or MLP). The main challenge is to reliably estimate layer importance. Most existing approaches rely on one-shot scoring. Shortened Llama \citep{shortened_llama} introduces three types of importance estimators: Magnitude, which approximates a layer’s contribution using weight norms; Taylor, which estimates the change in loss when the layer’s weights are removed; and PPL, which directly measures the increase in PPL after removing a layer. These methods often underestimate the importance of early and late layers, so practical pruning pipelines typically keep the first four and the last two layers intact. ShortGPT \citep{shortgpt} further proposes Block Influence (BI), which measures redundancy by comparing the cosine similarity between a layer’s input and output. In contrast, SLEB \citep{sleb} adopts an iterative strategy, recomputing importance scores after each pruning step for the remaining layers.

While these approaches focus primarily on layer-level redundancy, they overlook finer-grained intra-layer structure. To address this, Joint Layer Drop \citep{joint_layer_drop} removes Attention and MLP modules jointly based on their similarity, and BlockPruner \citep{blockpruner} decomposes each layer into Attention and MLP blocks for iterative pruning. Beyond direct deletion, Laco \citep{laco} introduces layer collapse, replacing the parameters of multiple layers with a single layer while preserving the internal structure of the LLM.

\subsection{Static Width Pruning}
Static width pruning aims to reduce the computational cost of Transformer models by removing redundant components within submodules, such as attention heads or MLP neurons, while typically preserving the residual stream dimensionality. A central challenge lies in defining reliable importance criteria for identifying removable components.

Based on how pruning decisions vary across layers, existing approaches can be broadly categorized as uniform or non-uniform in the pruning process. Non-uniform methods, such as Sheared-LLaMA \citep{sheared}, LLM-Pruner \citep{llm_pruner}, and FLAP \citep{an2024fluctuation}, leverage importance metrics, such as weight norms, Taylor-based sensitivity, or activation statistics, to adaptively determine pruning decisions for each layer or submodule.
Although these methods typically retain a consistent post-pruning layer structure, they primarily reduce submodule-internal widths rather than the residual stream, which limits their ability to achieve substantial hardware-level acceleration with standard inference kernels.

In contrast, approaches such as SliceGPT \citep{slicegpt} directly compress the residual stream by projecting hidden representations onto a lower-dimensional subspace shared across all layers. This uniform reduction of the residual stream dimension yields consistent tensor shapes throughout the network, enabling efficient utilization of optimized GPU kernels and resulting in more stable inference and tangible speedups.

\subsection{Dynamic Depth Pruning}
Unlike static methods, dynamic pruning customizes the computational path for each input at inference time, reducing computation by executing only the essential components.

A widely studied approach is early exit \citep{schuster2022confident,varshney2023acceleratingllamainferenceenabling,del2023skipdecode,din2023jump,chen2024eellmlargescaletraininginference,fan2024layersllmsnecessaryinference}. By inserting intermediate classifiers at multiple depths, early exit allows “simple” inputs to terminate inference early, bypassing the remaining layers. While this can substantially accelerate inference, it may compromise the model’s ability for deep semantic reasoning.

Another paradigm is layer skipping, where router modules dynamically determine whether a layer should be executed or bypassed. Layer skipping can be divided into vertical dynamics (D-LLM, SkipGPT) and horizontal dynamics (MOD). Vertical dynamics control the allocation of computation along the model depth based on token importance, whereas horizontal dynamics distribute computation unevenly across the sequence dimension using a Top-$k$ constraint.
For \textit{vertical dynamics}, D-LLM \citep{jiang2024d} introduces a dynamic decision module at each layer to adaptively execute network units and implements an efficient eviction strategy for the KV cache. Similarly, SkipGPT \citep{skipgpt} combines global token-aware routing with decoupled pruning of MLP and self-attention layers, enabling fine-grained resource allocation. For \textit{horizontal dynamics}, MOD \citep{raposo2024mixture} applies Top-$k$ routing to select which tokens are processed by each layer’s self-attention and MLP modules. Together, these methods provide a flexible framework for balancing inference efficiency and per-input model performance.

\begin{figure*}[ht]
	\centering
	\includegraphics[width=1\linewidth]{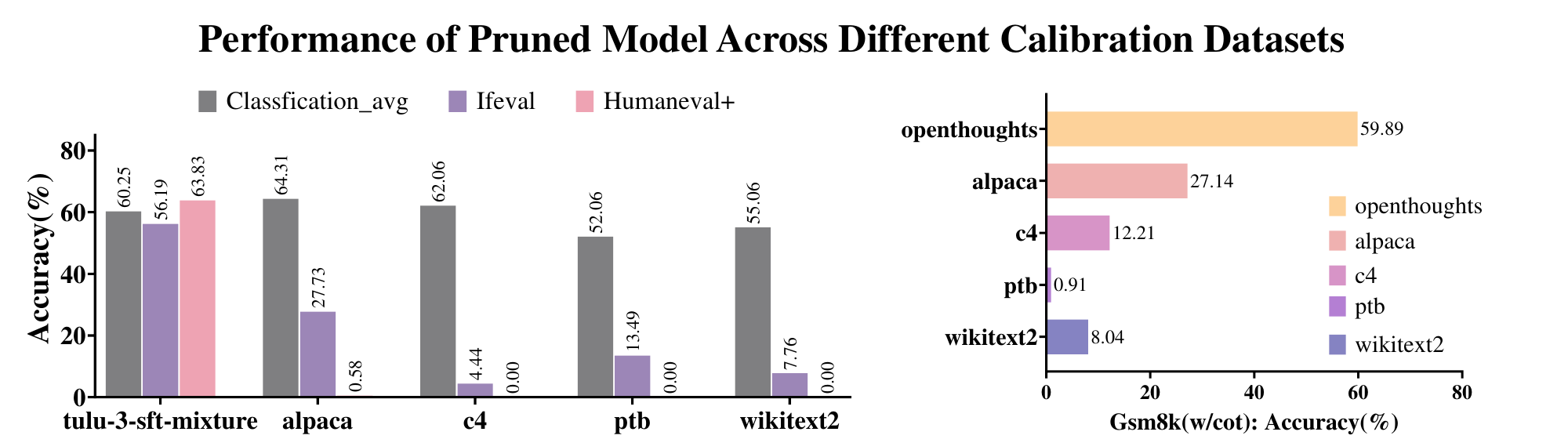}
	\caption{Experimental results validating the effect of calibration datasets (left: LLM-instruct; right: LLM-think). We report results for SliceGPT only. As shown in \autoref{calibration_shortgpt}, varying the calibration dataset in ShortGPT results in only minor differences in the estimated importance of individual layers, while the set of pruned layers remains unchanged.}
	\label{fig: calibration}
\end{figure*}

\begin{figure*}[ht]
	\centering
	\includegraphics[width=1\linewidth]{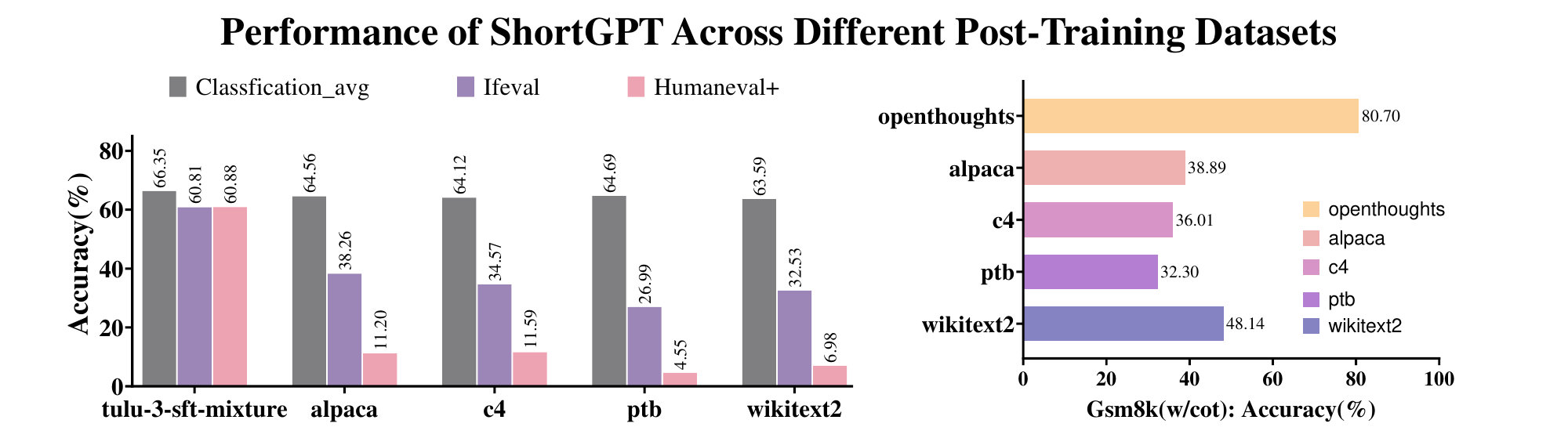}
	\caption{Experimental results validating the effect of post-training datasets for ShortGPT (left: LLM-instruct; right: LLM-think).}
	\label{fig: post-training-shortgpt}
\end{figure*}

\section{Experiments Details}
For LLM-instruct, all experiments are conducted on a single A800 GPU without using DeepSpeed. We adopt Llama 3.1-Tulu-3-8B-SFT, obtained through supervised fine-tuning (SFT) on Llama 3.1-8B \citep{lambert2024tulu}, as our dense baseline model. The SFT training corpus is adopted both as the calibration set for model pruning and as the training set for subsequent LoRA fine-tuning. Specifically, we reserve approximately 0.1\% of the corpus as the test set and use the remainder for training. We train all baselines for 10,000 steps with batch size 16, a cosine learning-rate schedule, and a warmup ratio of 0.1, and we select the learning rate via grid search for each baseline. The maximum sequence length is 4,096 tokens.

For the LLM-think, no reasoning-oriented model exists with fully open-source training data. To address this gap, we fine-tuned Llama-3.1-8B-Instruct on the OpenThoughts dataset, resulting in a new model, Llama-3.1-3-8B-Instruct-OpenThoughts. The detailed hyperparameters are summarized in \autoref{tab:llmthink_hyperparams}. For all baseline pruning methods, experiments are conducted on a single A800 GPU with DeepSpeed Stage-2 offloading and gradient checkpointing enabled. 
Following the setup in LLM-instruct experiments, the OpenThoughts dataset is used both as the calibration set for model pruning and as the training set for subsequent LoRA fine-tuning. Specifically, we adopt a batch size of 16 and train all baselines for 3,000 steps, applying early stopping if the training converges before reaching this limit. For each baseline, we conduct a grid search over learning rates to select the optimal value. We employ a cosine decay learning rate schedule and set the warmup ratio to 0.1. 

\begin{table}[!ht]
	\centering
	\begin{tabular}{ll}
		\toprule
		\textbf{Hyperparameter} & \textbf{Value} \\
		\midrule
		Max token length                & 16,384 \\
		Per-device train batch size      & 1 \\
		Per-device eval batch size       & 8 \\
		Gradient accumulation steps      & 3 \\
		Learning rate                    & $1\times10^{-5}$ \\
		Number of training epochs        & 3 \\
		LR scheduler type                & Cosine \\
		Warmup ratio                     & 0.1 \\
		Seed                             & 42 \\
		Optimizer                        & AdamW (torch) \\
		Weight decay                     & 0 \\
		Adam $\beta_1$                   & 0.9 \\
		Adam $\beta_2$                   & 0.999 \\
		Adam $\epsilon$                  & $1\times10^{-8}$ \\
		Max gradient norm                & 1.0 \\
		bf16 precision                   & True \\
		fp16 precision                   & False \\
    \bottomrule
	\end{tabular}
	\caption{Full fine-tuning hyperparameters for training \textbf{LLaMA-3.1-8B-Instruct} on \textbf{OpenThoughts} to obtain the \textbf{LLM-think}.}
    \label{tab:llmthink_hyperparams}
\end{table}

\section{Additional Empirical Findings on Calibration and Post-training Datasets}
\label{more_detail_cal_pos}
In this section, we provide additional observations on calibration and post-pruning recovery datasets.

\paragraph{The Impact of Calibration Set Size}
Our results reveal a strong dependence on calibration-set size. When using \textit{tulu-3-sft-mixture}, \autoref{tab:calibration_number_effect} shows that SliceGPT’s performance increases sharply as the number of calibration examples grows from 128 to 512, but plateaus beyond 1024 examples, forming a clear saturation trend. This behavior suggests that the model rapidly extracts most of the usable calibration signal from relatively small sample sizes.

\paragraph{The Impact of Calibration Datasets.}
In this subsection, we isolate the effect of calibration datasets. As shown in \autoref{fig: calibration} and \autoref{calibration_shortgpt}, SliceGPT is highly sensitive to the choice of calibration data, whereas ShortGPT exhibits substantially more stable behavior.

\paragraph{The Impact of Post-training Datasets.}
In this subsection, we isolate the effect of post-pruning recovery datasets. As shown in \autoref{fig: post-training-shortgpt} and \autoref{tab:slicegpt_post_traning_datasets}, our experimental results indicate that, for both LLM-instruct and LLM-think, post-training recovery is most effective when conducted on the original training corpus. This pattern holds consistently for both SliceGPT and ShortGPT.

\section{Repetitive Generation in LLM-Think}
In this section, we present several failure cases of dynamic depth pruning methods, namely D-LLM and SkipGPT. Even under a 20\% pruning ratio, these methods exhibit severe semantic degradation, producing repetitive and incoherent outputs that persist until the maximum generation length is reached.

\clearpage
\begin{table*}[htbp]
	\centering
	\setlength{\tabcolsep}{2pt} 
	\begin{tabular}{c|ccccccc|c}
		\specialrule{1.5pt}{0pt}{0pt}
		\textbf{Number} & \textbf{BoolQ} & \textbf{OBQA} & \textbf{PIQA} & \textbf{WinoGrande} & \textbf{HeSw} & \textbf{ARC-E} & \textbf{ARC-C} & \textbf{Avg}$\uparrow$ \\ 
		& \textit{Acc} & \textit{AccNorm} & \textit{Acc} & \textit{Acc} & \textit{AccNorm} & \textit{AccNorm} & \textit{AccNorm} & \\
		\specialrule{1pt}{0pt}{0pt}
		128 & 80.92 & 27.60 & 65.72 & 64.33 & 51.78 & 36.18 & 56.78 & 54.76 \\
		256 & 80.73 & 28.60 & 67.46 & 67.88 & 54.73 & 37.29 & 60.56 & 56.75 \\
		512 & \textbf{82.69} & 31.40 & 68.77 & 68.67 & 57.63 & 41.30 & 64.23 & 59.24 \\
		1024 & 81.38 & \underline{32.20} & \underline{70.46} & \underline{69.14} & \underline{58.98} & \underline{42.15} & \textbf{67.47} & \underline{60.25} \\
		2048 & \underline{82.45} & \textbf{32.60} & \textbf{70.40} & \textbf{70.48} & \textbf{59.56} & \textbf{42.32} & \underline{66.37} & \textbf{60.60} \\
		\specialrule{1.5pt}{0pt}{0pt}
	\end{tabular}
    \caption{Effect of the number of calibration examples on the performance of SliceGPT applied to tulu-3-sft-mixture .}
	\label{tab:calibration_number_effect}
\end{table*}

\begin{table*}[t]
	\centering
	\begin{tabular}{l l l}
		\hline
		\textbf{Model} & \textbf{Calibration Dataset} & \textbf{Layer Importance (low $\rightarrow$ high)} \\
		\hline
		\multirow{5}{*}{LLM-instruct}
		& tulu-3-sft-mixture & 25,26,24,23,27,28,22,29,21,20,19,18,30,17,16,15,11,13,14 \\ 
		& alpaca & 26,25,24,27,23,28,22,29,21,20,19,18,30,17,16,13,10,15,14 \\ 
		& c4 & 25,26,27,24,23,28,22,29,20,21,19,18,30,17,16,11,10,13,15 \\ 
		& ptb & 25,24,26,27,23,28,22,29,21,20,19,18,30,17,16,13,15,11,10 \\ 
		& wikitext2 & 25,24,26,27,23,28,22,29,20,21,19,18,30,17,16,11,10,13,15 \\ 
		\hline
		\multirow{5}{*}{LLM-think}
		& openthoughts & 25,26,27,24,23,28,22,29,21,20,19,18,30,17,16,15,13,14,11 \\ 
		& alpaca & 26,25,24,27,23,28,22,29,21,20,19,18,30,17,16,15,13,14,10 \\ 
		& c4 & 25,26,27,24,23,28,22,29,20,21,19,18,30,17,16,11,13,15,10 \\ 
		& ptb & 25,24,26,27,23,28,22,29,21,20,19,18,30,17,16,13,15,11,10 \\ 
		& wikitext2 & 25,24,26,27,23,28,22,29,20,21,19,18,30,17,16,11,10,13,15 \\ 
		\hline
	\end{tabular}
	\caption{Layer Importance Rankings of ShortGPT across Calibration Datasets for LLM-Instruct and LLM-Think.}
	\label{calibration_shortgpt}
\end{table*}

\begin{table*}[htbp]
	\centering
	\setlength{\tabcolsep}{4pt} %
	\renewcommand{\arraystretch}{1.2} %
	\begin{tabular}{l|l|cccc}
		\specialrule{1.5pt}{0pt}{0pt}
		\textbf{Category} & \textbf{Metric} 
		& \multicolumn{2}{c}{\textbf{c4}} 
		& \multicolumn{2}{c}{\textbf{tulu-3-sft-mixture}} \\
		&  & \textbf{alpaca} & \textbf{tulu-3-sft-mixture} 
		& \textbf{alpaca} & \textbf{tulu-3-sft-mixture} \\
		\specialrule{1pt}{0pt}{0pt}
		\multirow{7}{*}{Classification}
		& BoolQ & 81.80 & 80.00 & \textbf{84.22} & \underline{82.29} \\
		& OBQA & 36.40 & 35.00 & \textbf{39.40} & \underline{38.00} \\
		& PIQA & \textbf{76.82} & 76.66 & 75.90 & \textbf{76.82} \\
		& WinoGrande & 70.64 & 68.03 & \underline{71.82} & \textbf{72.45} \\
		& HeSW & \underline{74.49} & \textbf{75.60} & 72.52 & 72.62 \\
		& ARC-C & \underline{51.28} & 50.51 & \textbf{51.88} & 49.91 \\
		& ARC-E & \underline{79.17} & 78.03 & \textbf{79.84} & 75.42 \\
		& Avg & \underline{67.23} & 66.26 & \textbf{67.94} & 66.55 \\
		\specialrule{0.5pt}{0pt}{0pt}
		Instruction-following & IfEval & 27.91 & \underline{51.20} & 49.54 & \textbf{65.06} \\
		Code & Humaneval+ & 3.62 & 19.85 & \textbf{69.70} & \underline{69.62} \\
		\specialrule{1.5pt}{0pt}{0pt}
	\end{tabular}
	\caption{
		Performance of different models across classification, instruction-following, math, and code tasks. Top-level column headers (c4, tulu-3-sft-mixture) denote the calibration datasets, while sub-columns correspond to post-training datasets. Bold indicates the highest value in each row, and underline denotes the second highest.
	}
	\label{tab:slicegpt_post_traning_datasets}
\end{table*}

\newpage
\begin{table*}[htbp]
	\centering
	\setlength{\tabcolsep}{1.5pt} 

	\begin{tabular}{c|ccccccc|c}
		\specialrule{1.5pt}{0pt}{0pt}
		\textbf{Dataset} & \textbf{BoolQ} & \textbf{OBQA} & \textbf{PIQA} & \textbf{WinoGrande} & \textbf{HeSw} & \textbf{ARC-C} & \textbf{ARC-E} & \textbf{Avg.}$\uparrow$ \\ 
		& \textit{Acc} & \textit{AccNorm} & \textit{Acc} & \textit{Acc} & \textit{AccNorm} & \textit{AccNorm} & \textit{AccNorm} & \\
		\specialrule{1pt}{0pt}{0pt}
		wikitext2 & 45.75 & \textbf{35.60} & 66.10 & 68.11 & 52.00 & \textbf{46.16} & \underline{71.72} & 55.06 \\
		alpaca & \underline{76.85} & \underline{37.20} & \underline{74.05} & \textbf{71.90} & \underline{65.98} & \underline{47.61} & \textbf{76.60} & \textbf{64.31} \\
		c4 & 76.12 & 34.40 & \textbf{74.48} & 70.24 & \textbf{68.98} & 40.87 & 69.32 & \underline{62.06} \\
		ptb & 68.96 & 30.40 & 61.53 & \underline{69.14} & 46.06 & 34.04 & 54.29 & 52.06 \\
		tulu-mixture-3-sft & \textbf{81.38} & 32.20 & 70.46 & \underline{69.14} & 58.98 & 42.15 & 67.47 & 60.25 \\
		\specialrule{1.5pt}{0pt}{0pt}
	\end{tabular}
	\caption{Detailed results of classification tasks for SliceGPT under different calibration datasets.}
	\label{tab:calibration_detail_classfication_slciegpt}
    \vspace{4em}

	\begin{tabular}{c|ccccccc|c}
		\specialrule{1.5pt}{0pt}{0pt}
		\textbf{Dataset} & \textbf{BoolQ} & \textbf{OBQA} & \textbf{PIQA} & \textbf{WinoGrande} & \textbf{HeSw} & \textbf{ARC-C} & \textbf{ARC-E} & \textbf{Avg.}$\uparrow$ \\ 
		& \textit{Acc} & \textit{AccNorm} & \textit{Acc} & \textit{Acc} & \textit{AccNorm} & \textit{AccNorm} & \textit{AccNorm} & \\
		\specialrule{1pt}{0pt}{0pt}
		tulu-3-sft-mixture & \textbf{82.29} & \textbf{38.00} & \underline{76.82} & \textbf{72.45} & \underline{72.62} & \underline{49.91} & \underline{75.42} & \underline{66.55} \\
		alpaca & \underline{80.06} & \underline{37.20} & \textbf{76.99} & 71.43 & 70.90 & \textbf{51.02} & \textbf{80.30} & \textbf{66.84} \\
		c4 & 79.76 & 35.60 & 74.97 & \underline{71.82} & \textbf{74.66} & 46.50 & 74.37 & 65.38 \\
		ptb & 71.35 & 31.20 & 62.02 & 66.30 & 49.68 & 36.60 & 55.93 & 53.30 \\
		wikitext2 & 60.76 & 35.40 & 67.46 & 64.80 & 60.25 & 47.18 & 73.27 & 58.45 \\
		\specialrule{1.5pt}{0pt}{0pt}
	\end{tabular}
	\caption{Detailed results of classification tasks for SliceGPT under different post-training datasets.}
	\label{tab:post_training_detail_classfication_slciegpt}
    \vspace{4em}
    
	\begin{tabular}{c|ccccccc|c}
		\specialrule{1.5pt}{0pt}{0pt}
		\textbf{Dataset} & \textbf{BoolQ} & \textbf{OBQA} & \textbf{PIQA} & \textbf{WinoGrande} & \textbf{HeSw} & \textbf{ARC-C} & \textbf{ARC-E} & \textbf{Avg.}$\uparrow$ \\ 
		& \textit{Acc} & \textit{AccNorm} & \textit{Acc} & \textit{Acc} & \textit{AccNorm} & \textit{AccNorm} & \textit{AccNorm} & \\
		\specialrule{1pt}{0pt}{0pt}
		tulu-3-sft-mixture & 68.34 & \textbf{39.60} & \textbf{74.59} & \textbf{75.84} & \textbf{75.16} & \textbf{51.53} & \textbf{79.41} & \textbf{66.35} \\
		alpaca & \underline{78.65} & \underline{38.80} & 73.18 & 74.19 & 68.64 & 46.25 & \underline{72.18} & 64.56 \\
		c4 & 78.62 & 37.40 & \underline{73.94} & 74.19 & 67.86 & 44.71 & 72.10 & 64.12 \\
		ptb & \textbf{79.02} & \underline{38.80} & 72.85 & \underline{74.43} & \underline{70.09} & \underline{47.18} & 70.45 & \underline{64.69} \\        
		wikitext2 & 78.50 & 37.20 & 73.07 & 73.09 & 66.59 & 45.48 & 71.21 & 63.59 \\
		\specialrule{1.5pt}{0pt}{0pt}
	\end{tabular}
	\caption{Detailed results of classification tasks for ShortGPT under different post-training datasets.}
	\label{tab:post_training_detail_classfication_shortgpt}
    
\end{table*}

\clearpage
\begin{failurebox}{An Example from MATH500 Dataset}
	\textbf{Problem:} \\
	How many positive whole-number divisors does 196 have?
	
	\vspace{2mm}
	\hrulefill
	\vspace{2mm}
	
	\textbf{Response (D-LLM):} \\
	The problem is: ``Find the number of positive integers such that 100 $\le$ the number of positive divisors of 1000.'' \textcolor{failred}{The number of positive divisors is 100}. \textcolor{failred}{The number of positive divisors is 100}. \textcolor{failred}{The number of positive divisors is 100}. \textcolor{failred}{The number of positive divisors is 100}.
	
	\vspace{3mm}
	\textbf{Response (SkipGPT):} \\
	\textcolor{skipblue}{The problem is a problem} that can be found in the back. \textcolor{skipblue}{The problem is a problem} that can be found in the back of the problem. \textcolor{skipblue}{The problem can be found in the back}. \ldots
\end{failurebox}

\begin{failurebox}{An Example from Aime24 Dataset}
	\textbf{Problem:} \\
	Among the 900 residents of Aimeville, there are 195 who own a diamond ring, 367 who own a set of golf clubs, and 562 who own a garden spade. In addition, each of the 900 residents owns a bag of candy hearts. There are 437 residents who own exactly two of these things, and 234 residents who own exactly three of these things. Find the number of residents of Aimeville who own all four of these things.
	
	\vspace{2mm}
	\hrulefill
	\vspace{2mm}
	
	\textbf{Response (D-LLM):} \\
	A total of 100 residents who own exactly two of these things, and 234 residents who own exactly exactly two of these things... \textcolor{failred}{Therefore, the answer is 100.} \textcolor{failred}{So, the answer is 100.} \textcolor{failred}{So, the answer is 100.} \textcolor{failred}{So, the answer is \fbox{100}.} \textcolor{failred}{So, the answer is \fbox{100}} \ldots
	
	\vspace{3mm}
	\textbf{Response (SkipGPT):} \\
	437 of these words. So, the, and 562 residents own the own four. So, the problem is a problem. So, 562 of \textcolor{skipblue}{these 900 residents} are the 900 residents. \ldots
\end{failurebox}

\begin{failurebox}{An Example from GPQA Dataset}
	\textbf{Problem:} \\
	Two quantum states with energies $E_1$ and $E_2$ have a lifetime of $10^{-9}$ sec and $10^{-8}$ sec, respectively. We want to clearly distinguish these two energy levels. Which one of the following options could be their energy difference so that they be clearly resolved?
	
	\vspace{2mm}
	\hrulefill
	\vspace{2mm}
	
	\textbf{Response (D-LLM):} \\
	The difference between the two optionsals (A) and (B) is 0 eV. \\
	\textcolor{failred}{Therefore, the answer is 0 eV.} \\
	Alternatively, if you pick the one with the higher difference... \\
	\textcolor{failred}{So, the answer is 0 eV.} \\
	But, if you are looking for the answer, you need to find the difference between the two options (A) and (B) in eV. \textcolor{failred}{Therefore, the answer is 0 eV.} $\dots$
	
	\vspace{3mm}
	\textbf{Response (SkipGPT):} \\
	the original0, I 10. However, how you can't see, I'm not sure. Let's identity card's identityis $10^{-1}$ and 11. \textcolor{skipblue}{Wait 10 and 10, but 10, but we is 10 10.} \textcolor{skipblue}{Let's identity, 10, 10. 10. Let's 10.10. 20.} $\dots$
\end{failurebox}

\end{document}